\documentclass[10pt,twocolumn,letterpaper]{article}

\usepackage{cvpr}              

\usepackage[dvipsnames]{xcolor}

\definecolor{cvprblue}{rgb}{0.21,0.49,0.74}
\usepackage[pagebackref,breaklinks,colorlinks,citecolor=cvprblue]{hyperref}
\usepackage{colortbl}
\usepackage{bbding}
\usepackage{graphicx}
\usepackage{amsmath}
\usepackage{amssymb}
\usepackage{booktabs}
\usepackage{graphics}
\usepackage{epstopdf}
\usepackage{color}
\usepackage{multirow}
\usepackage[accsupp]{axessibility}
\usepackage[capitalize]{cleveref}
\crefname{section}{Sec.}{Secs.}
\Crefname{section}{Section}{Sections}
\Crefname{table}{Table}{Tables}
\crefname{table}{Tab.}{Tabs.}
\newcommand{\aka}{\textit{a.k.a.}}

\def\paperID{3645} 
\def\confName{CVPR}
\def\confYear{2024}

\title{Elite360D: Towards Efficient 360 Depth Estimation via Semantic- and Distance-Aware Bi-Projection Fusion}

\author{Hao Ai$^{1}$ \quad Lin Wang$^{1}$$^{,2}$\thanks{Corresponding author (e-mail: linwang@ust.hk)}\\
$^{1}$AI Thrust, HKUST(GZ) \quad $^{2}$Dept. of CSE, HKUST\\
{\tt\small hai033@connect.hkust-gz.edu.cn, linwang@ust.hk}}

\begin{document}
\maketitle
\begin{abstract}
360 depth estimation has recently received great attention for 3D reconstruction owing to its omnidirectional field of view (FoV).
Recent approaches are predominantly focused on cross-projection fusion with geometry-based re-projection: they fuse 360 images with equirectangular projection (ERP) and another projection type, \eg, cubemap projection, to estimate depth with the ERP format. However, these methods suffer from 1) limited local receptive fields, making it hardly possible to capture large FoV scenes, and 2) prohibitive computational cost, caused by the complex cross-projection fusion module design. In this paper, we propose \textbf{\textit{Elite360D}}, a novel framework that inputs the ERP image and icosahedron projection (ICOSAP) point set, which is undistorted and spatially continuous. Elite360D is superior in its capacity in learning a representation from a local-with-global perspective. With a flexible ERP image encoder, it includes an ICOSAP point encoder, and a Bi-projection Bi-attention Fusion (B2F) module (totally \textit{$\thicksim$1M parameters}). Specifically, the ERP image encoder can take various perspective image-trained backbones (\eg, ResNet, Transformer) to extract local features. The point encoder extracts the global features from the ICOSAP. Then, the B2F module captures the semantic- and distance-aware dependencies between each pixel of the ERP feature and the entire ICOSAP feature set. Without specific backbone design and obvious computational cost increase, Elite360D outperforms the prior arts on several benchmark datasets.
\end{abstract}
\vspace{-16pt}
\section*{Multimedia Material}
\vspace{-5pt}
For videos, code, demo and more information, you can visit \href{https://VLIS2022.github.io/Elite360D/}{https://VLIS2022.github.io/Elite360D/}    
\vspace{-8pt}
\section{Introduction}
\label{sec:intro}
\begin{figure}[!t]
\centering
\includegraphics[width=0.85\linewidth]{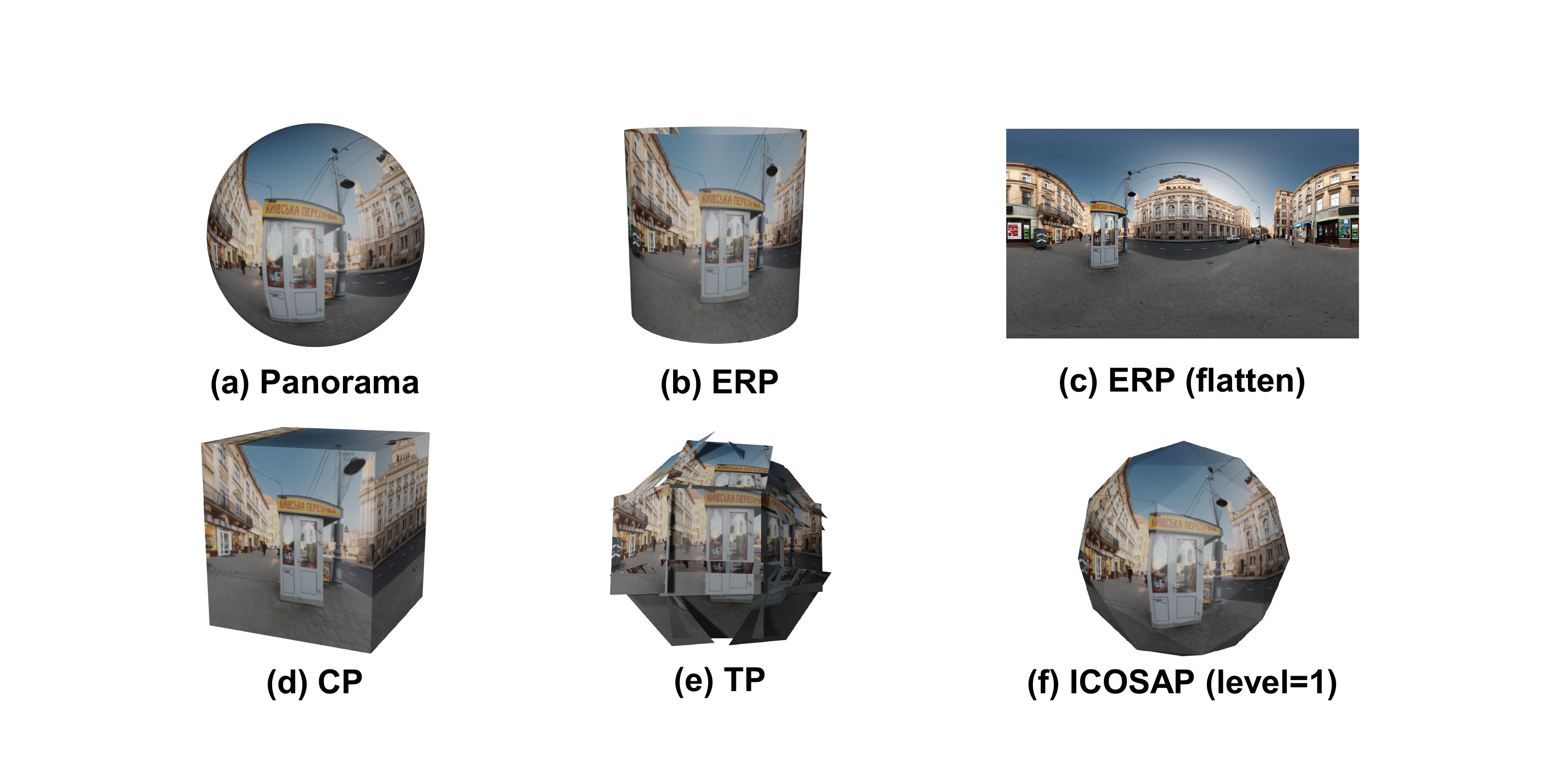}
\vspace{-8pt}
\caption{Different projections of a spherical imaging panorama.
}
\vspace{-15pt}
 \label{fig:representation}
\end{figure}

\vspace{-3pt}
360$^\circ$ images capture the complete surrounding environment in one shot with an ultra-wide field-of-view (FoV) of 180$^\circ$$\times$360$^\circ$, which is broadly applied to applications, \eg, autonomous driving~\cite{RaviKumar2021OmniDetSV,Yoon2021SphereSRI, Ye2022Rope3DTR, 2022DeepLFAi,Zheng2023BothSA} and virtual reality ~\cite{Martin2021ScanGAN360AG,Li2021LookingHO,Cao2023OmniZoomerLT}. The ability to infer the 3D structure of a surrounding scene has inspired active research for monocular 360 depth estimation.
Equirectangular projection (ERP) is the commonly used projection type that can provide a complete view of a scene. An ERP image, \aka, panorama, samples pixels with a higher density at the poles compared to the equator, resulting in spherical distortions (Fig.~\ref{fig:representation}(c)).



 
\begin{figure}[!t]
\centering
\includegraphics[width=\linewidth]{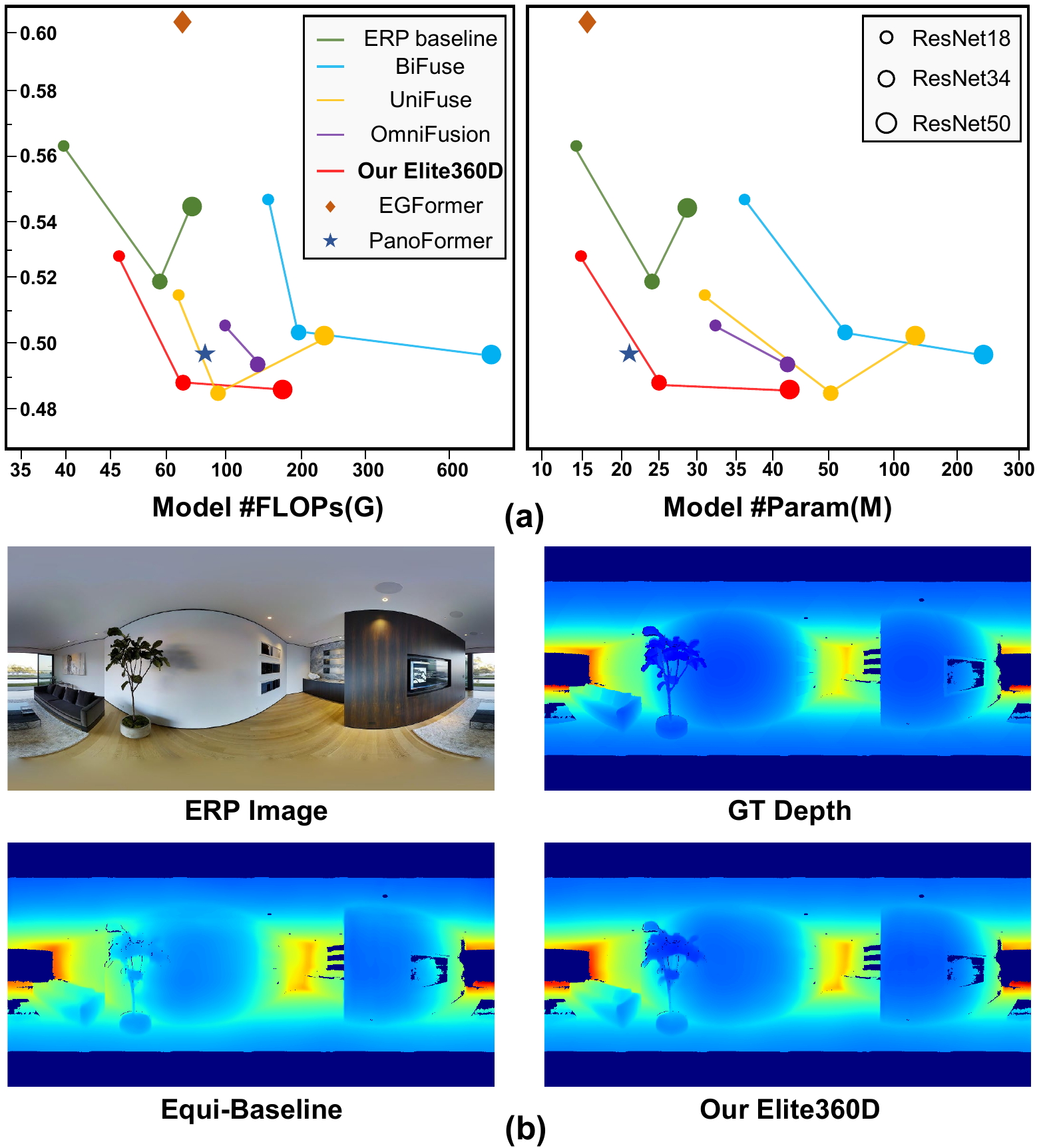}
\vspace{-20pt}
\caption{(a) Performance (RMSE error) curves on M3D test dataset~\cite{Chang2017Matterport3DLF}. Larger circles mean more parameters (\eg, ResNet18, ResNet34, ResNet50) and lower errors mean better performance. (b) Comparison with the ResNet34 as the ERP encoder backbone. With only 1M more parameters, our depth result is more accurate.}
\vspace{-15pt}
 \label{fig:coverfig}
\end{figure}

In recent years, to address the spherical distortions, several ah-hoc designs~\cite{Zioulis2018OmniDepthDD, Tateno2018DistortionAwareCF,Su2018KernelTN,Zhuang2021ACDNetAC} focus on the distortion-aware convolutional kernels and sample grids from undistorted tangent planes on the sphere. These grids are back-projected to the corresponding locations of the ERP image. Also, some methods~\cite{Pintore2021SliceNetDD,Sun2020HoHoNet3I,Yu2023PanelNetU3} partition the ERP input into multiple vertical slices and extract a compact representation from these slices to extend the receptive field of each pixel for the ERP representation. Unfortunately, these methods result in substantial computational costs. Meanwhile, the limited local receptive field of the convolutional filters is insufficient to provide the global perception of the panorama. Therefore, PanoFormer~\cite{Shen2022PanoFormerPT} and EGFormer~\cite{Yun2023EGformerEG} design specific transformer architectures to effectively model the long-range dependencies between the global context of large-FoV ERP images. Nevertheless, the receptive fields are confined by the sizes of local attention windows. Furthermore, the aforementioned data-driven methods often yield sub-optimal results when trained from scratch on limited 360$^\circ$ depth dataset (See Tab.~\ref{tab:comparison-to-soat}). 

With the power of existing large-scale perspective images~\cite{JiaDeng2009ImageNetAL}, several methods~\cite{Wang2020BiFuseM3, Jiang2021UniFuseUF,Wang2022BiFuseSA,Ai2023HRDFuseM3} employ the pre-trained models -- designed for perspective images -- as encoder backbones to extract ERP feature maps, and propose the cross-projection fusion to rectify distortions in the ERP feature maps. Specifically, these methods introduce the less-distorted planar projection data, \ie, cubemap projection (CP) patches or tangent projection (TP) patches (See Fig.~\ref{fig:representation}(d), (e)), and unify the spatial dimensions between different projections for the cross-projection fusion. BiFuse and UniFuse~\cite{Jiang2021UniFuseUF,Wang2020BiFuseM3,Wang2022BiFuseSA} propose the C2E module to re-project 
the content in patch-wise CP features into ERP grid following the spherical geometric relationships. Furthermore, they leverage the fusion between ERP feature maps and C2E feature maps to enhance depth estimation accuracy. However, their concatenation-based feature fusion across multiple scales brings a significant computational burden. Based on the feature similarity, HRDFuse~\cite{Ai2023HRDFuseM3} spatially aligns the ERP pixel features and TP patch features. This makes it possible to significantly marry the global context in the ERP feature map and regional structural details in the TP patch features. However, the decoder-level fusion remarkably increases the computational memory and cost.

These less-distorted gnomonic projections,~\ie, CP and TP patches, are spatially discontinuous and require complex re-projection operations for the desired ERP format predictions. Inspired by these issues, we introduce a more powerful projection, the icosahedron projection (ICOSAP)~\cite{Lee2018SpherePHDAC,Zhang2019OrientationAwareSS}, see Fig.~\ref{fig:representation}(f). Importantly, ICOSAP is a spatially continuous and globally perceptive non-Euclidean projection for 360$^\circ$ images.
In light of this, we propose an efficient and effective fusion-based framework, named \textbf{Elite360D} that takes the best of ERP and ICOSAP by learning a representation from a local-with-global perspective. 
It comprises three components: an ERP encoder, an ICOSAP encoder, and the bi-projection bi-attention fusion (B2F) module. For the ERP image encoder, our Elite360D flexibly supports a wide range of off-the-shelf 2D models, \eg, ResNet~\cite{He2015DeepRL}, Swin transformer~\cite{Liu2021SwinTH}, as the backbone to extract ERP feature maps. This potentially reduces the overfitting problems, particularly on small-scale 360 depth estimation datasets~\cite{Chang2017Matterport3DLF,Armeni2017Joint2D}. For efficiency, we represent the ICOSAP spheres as the discrete point sets rather than icosahedron meshes~\cite{Zhang2019OrientationAwareSS,Cohen2019GaugeEC,Jiang2019SCUG} or unfolded representations~\cite{Lee2018SpherePHDAC,Yoon2021SphereSRI}. As such, we can avoid semantic information redundancy due to dense ERP pixels and maintain the spatial position information. Then, to enable each ERP pixel feature with local receptive fields to perceive the whole scene, B2F module captures the semantic- and distance-aware dependencies between each ERP pixel feature and entire ICOSAP feature set.


We conduct extensive experiments on three different datasets with different encoder backbones to demonstrate the flexibility and effectiveness of our $\textit{Elite360D}$. The experimental results indicate that our Elite360D significantly improves plain-backbones' performance with minimal computational memory (\textit{only about 1M parameters}) (See Fig.~\ref{fig:coverfig}(a)).
Note that, based on the simple ERP depth estimation baseline, $\textit{Elite360D}$ only performs bi-projection feature fusion at the last feature layer and achieves results that are on par with leading methods, such as~\cite{Ai2023HRDFuseM3,Li2022OmniFusion3M}.

In summary, our main contributions are three-fold: (I) We introduce the ICOSAP with spatial continuity and global perception and represent them as the discrete point sets to reduce the computation cost and maintain the spatial information; (II) We propose the B2F module that jointly models the semantic and spatial dependencies between ICOSAP and ERP features and learn the representations with global-with-local receptive fields for large-FoV scenes; (III) Building upon the B2F module, we propose a novel framework that supports diverse off-the-shelf models as encoder backbones, showing better flexibility than the specially designed models, \eg,~\cite{Shen2022PanoFormerPT,Yun2023EGformerEG}.

\section{Related Works}
\label{sec:rela_work}
\noindent\textbf{Monocular 360 depth estimation}
Existing methods can be categorized into two types: one with single projection input, and the other with bi-projection inputs.

\begin{figure*}[t!]
\centering
\includegraphics[width=0.93\textwidth]{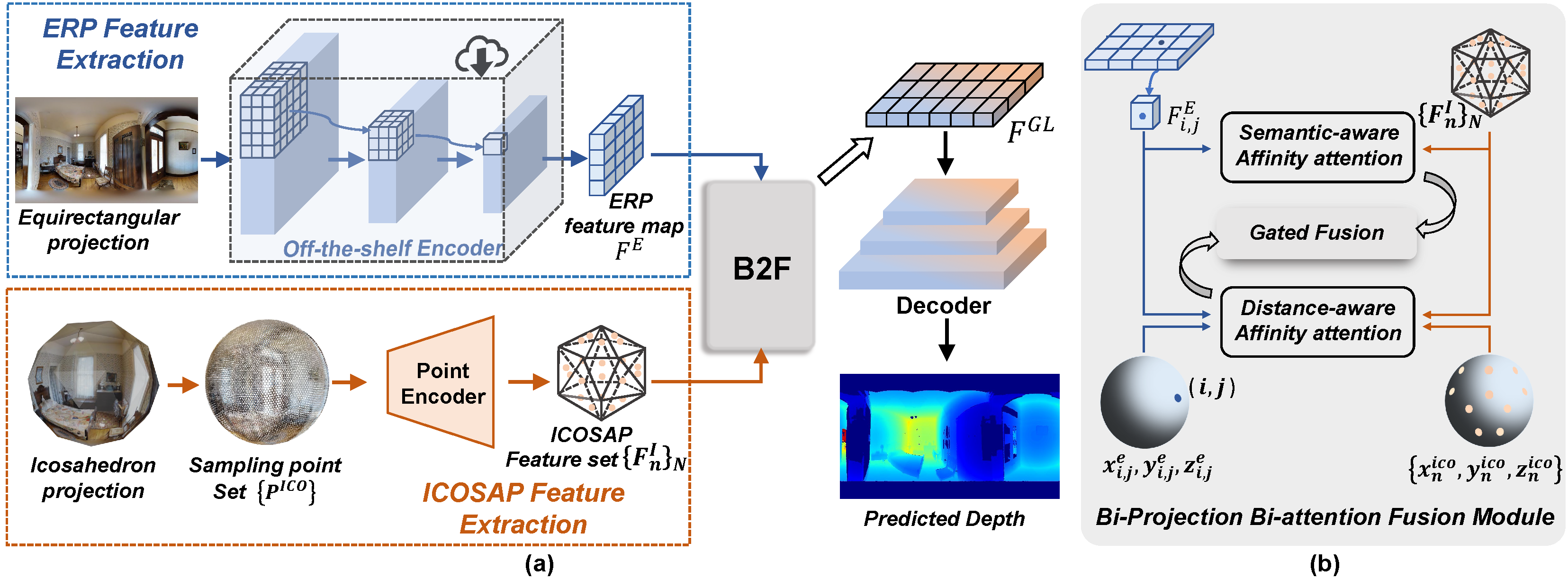}
\vspace{-8pt}
\caption{(a)\textbf{ An overview of our \textit{Elite360D framework}}, comprising image-based ERP feature extraction (Sec.~\ref{sec:erp_feature}), point-based ICOSAP feature extraction (Sec.~\ref{sec:ico_feature}), and Bi-projection Bi-attention fusion (B2F) (Sec.~\ref{sec:b2f}). For better visualization, we do not show the skip connections~\cite{Ronneberger2015UNetCN} at the decoding stage. (b) Illustration of the B2F module, consisting of three parts: semantic-aware affinity attention block (Fig.~\ref{fig:SAattenion}), distance-aware affinity attention block (Fig.~\ref{fig:DAattenion}) and gated fusion (Eq.~\ref{eq:gate_fusion}). }
\vspace{-15pt}
\label{fig:overview}
\end{figure*}

\textbf{1) Single Projection Input}
With the ERP images as the inputs, many methods focus on mitigating the inherent spherical distortion issues. Several methods~\cite{Tateno2018DistortionAwareCF,Cheng2020ODECNNOD,Zioulis2018OmniDepthDD} rectify distortions by modifying the receptive fields of traditional convolutional kernels.
Besides, OmniFusion~\cite{Li2022OmniFusion3M} and 360MonoDepth~\cite{ReyArea2022360MonoDepthH3} follow~\cite{Eder2019TangentIF} to predict depth maps from less distorted TP patches. Similarly, PanelNet~\cite{Yu2023PanelNetU3} predicts depth maps on the vertical slices of ERP images and then aggregates them. Moreover, some methods~\cite{Zhuang2021ACDNetAC,Pintore2021SliceNetDD,Sun2020HoHoNet3I} focus on tackling the small receptive fields of convolution filters for processing large FoV scenes. More recently, PanoFormer~\cite{Shen2022PanoFormerPT} and EGFormer~\cite{Yun2023EGformerEG} design transformer architectures~\cite{dosovitskiy2021an} for ERP images to model the long-range dependencies. By contrast, S$^2$Net~\cite{Li2023mathcalA} extracts robust features from ERP images with the Swin transformer~\cite{Liu2021SwinTH} and projects the ERP feature maps onto the spherical surface to minimize distortions and maintain spatial consistency.

\textbf{2) Bi-Projection Inputs.} BiFuse~\cite{Wang2020BiFuseM3} utilizes a dense fusion strategy to bidirectionally fuse ERP and CP features at both encoding and decoding stages. In contrast, BiFuse++\cite{Wang2022BiFuseSA} and UniFuse~\cite{Jiang2021UniFuseUF} fuse ERP and CP features at the encoding stage. Recently, HRDFuse~\cite{Ai2023HRDFuseM3} achieved the SOTA performance based on the adaptive fusion of ERP and TP predictions. By contrast, we introduce a non-Euclidean projection, ICOSAP, which is less distorted and spatially continuous. We leverage the global perception capacity of ICOSAP to enable each ERP pixel-wise feature, with a limited local receptive field, to capture the entire scene.

\begin{figure}[!t]
\centering
\includegraphics[width=\linewidth]{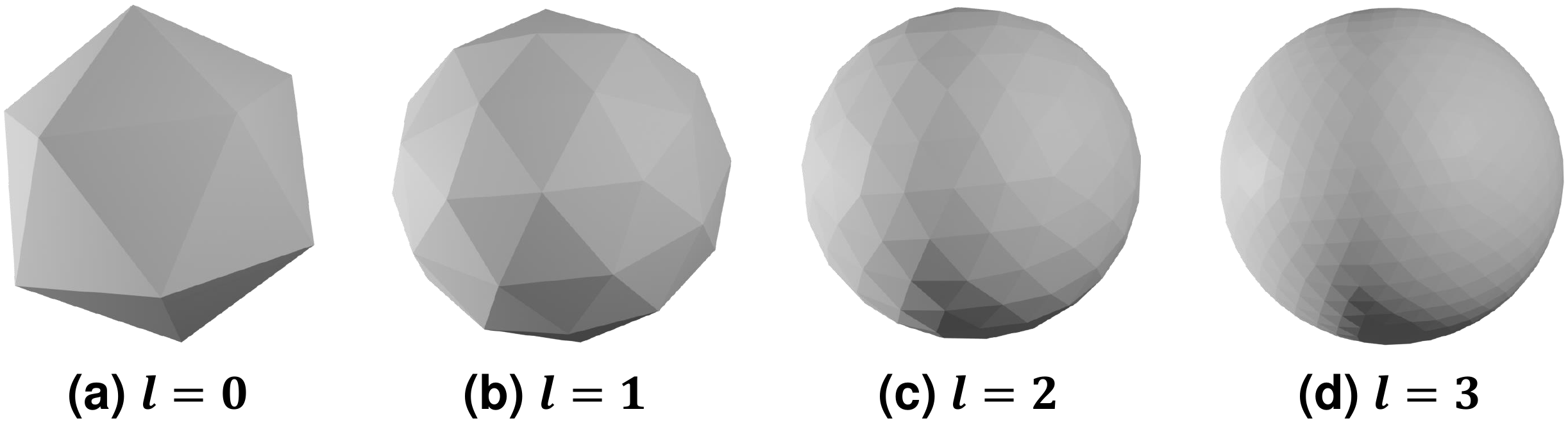}
\caption{ The subdivision of icosahedron at different resolution $l$.}
\vspace{-15pt}
 \label{fig:icosap}
\vspace{-5pt}
\end{figure}

\noindent \textbf{Representations for 360 Images}
ERP is the most commonly used projection, which projects a panorama onto a cylinder (Fig.~\ref{fig:representation}(b)) and then unfolds it into a plane (Fig.~\ref{fig:representation}(c)). However, ERP maps latitudes and longitudes onto the vertical and horizontal axes with equal spacing on the plane, leading to distortions. CP~\cite{Monroy2017SalNet360SM} projects the panoramas onto six faces of a cube where each face is less distorted (Fig.~\ref{fig:representation}(d)). 
Building upon CP, some padding methods~\cite{Cheng2018CubePF,Wang2020BiFuseM3} remove the boundary inconsistency between the adjacent CP patches. 
Recently, Eder \etal~\cite{Eder2019TangentIF} proposed the tangent projection (TP) (Fig.~\ref{fig:representation}(e)), to significantly mitigate spherical distortion. Besides,~\cite{Lee2018SpherePHDAC,Shakerinava2021EquivariantNF,Yoon2021SphereSRI} employ the polyhedrons, \eg, octahedron and icosahedron, to represent the panoramas and unfold these polyhedron representations on the plane for processing. 
In Elite360D, we select popular ICOSAP (See Fig.~\ref{fig:icosap}) as our input. Notably, we represent ICOSAP as discrete points, significantly reducing the computational cost while preserving spatial information and global awareness. Meanwhile, with the dense pixels in the ERP panoramas, discrete ICOSAP points can avoid semantic information redundancy.

\noindent \textbf{Cross-Attention Mechanism}
It is widely used for efficient multi-modal feature fusion. For instance, Chen~\etal~\cite{Chen2022AutoAlignPF} built cross-attention blocks to align and fuse 2D image features and 3D point cloud features for 3D object recognition. Similarly, BEVGuide~\cite{Man2023BEVGuidedMF}, using BEV embedding as a guided query, employs a cross-attention block to fuse information across different sensors. We propose the B2F module to model the relationships between each pixel of the ERP feature and the whole ICOSAP point feature set, based on the semantic-aware affinity attention and distance-aware affinity attention simultaneously. With the B2F module, each ERP pixel feature can perceive the spatial information and semantic information of the large FoV scenes.

\vspace{-18pt}
\section{The Proposed Framework}
\vspace{-4pt}
\noindent\textbf{Overview.}
Our goal is to investigate an efficient and effective bi-projection fusion framework for 360 depth estimation. To this end, as illustrated in Fig.~\ref{fig:overview}, the proposed Elite360D consists of three key components: image-based ERP feature extraction, point-based ICOSAP feature extraction and bi-projection bi-attention fusion (B2F). For the ERP feature $F^{E}$ extraction, Elite360D flexibly accommodates a wide range of 2D models as encoder backbones and takes advantage of pre-training to reduce the over-fitting problems. For the ICOSAP feature extraction, we represent ICOSAP sphere as the point set, which significantly reduces the computation cost and preserves the spatial information and global perception. Then we directly employ the simple point encoder, \ie Point transformer~\cite{Zhao2020PointT}, to extract the ICOSAP point feature set $\{F^{I}_n\}_{n=1,\cdots,N}$, $N$ is the point number, denoted as $\boldsymbol{\{F^{I}_n\}_N}$ . Next, to facilitate bi-projection feature fusion and learn a powerful representation $F^{GL}$ from a local-with-global perspective, B2F module comprises semantic-aware affinity attention block, distance-aware affinity attention block, and gated fusion block. These two attention blocks model the semantic- and distance-aware dependencies between each ERP pixel-wise feature and entire ICOSAP feature set, respectively and the gated fusion block adaptively fuses the bi-attention outputs. Lastly, we employ a simple decoder with up-sampling and skip-connections to predict the final depth map from the fused representation $F^{GL}$. We now describe the details.

\vspace{-3pt}
\subsection{ERP Feature Extraction}
\label{sec:erp_feature}
\vspace{-3pt}
Taking an ERP image with the resolution size of $H\times W$ as the input, the encoder extracts feature map $F^{E} \in \mathbb{R}^{h\times w\times C} $, where $h=H/s$, $w=W/s$, $s$ is the down-sampling scale factor and $C$ is the channel number. In particular, as treating ERP images as 2D perspective images, our encoder backbone is compatible with a wide range of robust 2D models that have been pre-trained on large-scale perspective image datasets~\cite{JiaDeng2009ImageNetAL}, including CNNs and vision transformers, \eg, ResNet~\cite{He2015DeepRL}, EfficientNet~\cite{Tan2019EfficientNetRM}, Swin transformer~\cite{Liu2021SwinTH}. Notably, as repeated local operations cannot substitute for a global operator~\cite{Wang2017NonlocalNN,Yun2023EGformerEG}, the finite size of convolutional kernels or the limited size of local attention windows results in ERP pixel-wise features lacking sufficient global receptive fields. Additionally, these 2D model performance suffers from the spherical distortions (See Tab.~\ref{tab:comparison-to-erp}).

\vspace{-3pt}
\subsection{ICOSAP Feature Extraction}
\label{sec:ico_feature}
\vspace{-3pt}
\noindent\textbf{Preliminary.} Since Lee~\etal~\cite{Lee2018SpherePHDAC} found that polyhedron representations with a greater number of initial faces exhibit lower distortion, this paper focuses on the ICOSAP with the most initial faces. Especially, we introduce the ICOSAP to offer a comprehensive and high-quality global perception. For the resolution, ICOSAP data consists of $20\times4^l$ faces and $12\times4^l$ vertices at the subdivision level $l$ (See Fig.~\ref{fig:icosap}). Given the spatial relationships between ICOSAP and ERP within the spherical space, RGB values of ICOSAP vertices can be derived from corresponding ERP pixels.

\noindent\textbf{Feature extraction.} Existing works focus on processing ICOSAP data on the plane, \ie, unfolded mesh representation~\cite{Zhang2019OrientationAwareSS} and spherical polyhedron representations~\cite{Lee2018SpherePHDAC}. However, these methods require specially designed and computationally intensive operations, such as convolutions, pooling~\cite{Lee2018SpherePHDAC}, and up-sampling~\cite{Shakerinava2021EquivariantNF}, to process the planar representations. By contrast, we propose to represent an ICOSAP input as a point set, as illustrated at the bottom of Fig.~\ref{fig:overview}(a). ERP images provide dense pixel values and discrete ICOSAP point set can prevent redundancy in semantic information while preserving spatial information and global perception. Therefore,  we take full advantage of two projections for efficient and accurate 360 depth estimation.
In detail, we employ the central points of each face to represent the ICOSAP sphere. Firstly, we obtain the $20\times4^l$ faces of an ICOSAP sphere at the default subdivision level $l$, where each face is composed of three vertices. Then, we calculate the spatial coordinates and RGB values of each face center by averaging them of each three vertices, as each face is an equilateral triangle. As a result, we obtain the ICOSCAP point set $\{\boldsymbol{P^{ICO}}\}\in \mathbb{R}^{(20\times4^l)\times6}$, where $20\times4^l$ is the point number and $6$ represents the coordinates [$x, y, z$] and RGB channels. With the input point set $\{\boldsymbol{P^{ICO}}\}$, we directly employ the encoder of Point Transformer~\cite{Zhao2020PointT} to extract ICOSAP point feature set $\boldsymbol{\{F^{I}_n\}_N}\in \mathbb{R}^{N\times C}$, where $N$ represents the number of point features and $C$ corresponds to the same channel number as the ERP feature map $F^{E}$. Especially, for simplicity and efficiency, we do not opt for complex networks for this purpose.
\begin{figure}[!t]
\centering
\includegraphics[width=0.90\linewidth]{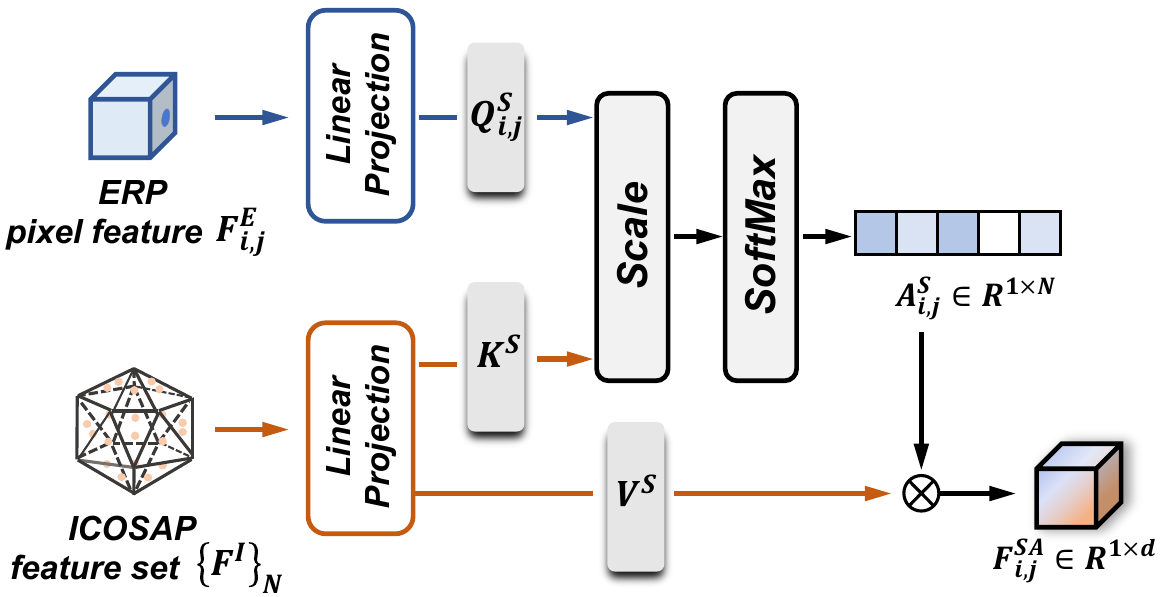}
\vspace{-8pt}
\caption{The architecture of semantic-aware affinity attention. Especially, $\mathbf{Q}^{S}_{i,j} \in \mathbb{R}^{1 \times d}$, $\mathbf{K}^{S} \in \mathbb{R}^{N \times d}$, and $\mathbf{V}^{S} \in \mathbb{R}^{N \times d}$, where $d$ is the dimension and $N$ is the ICOSAP point number.}
\label{fig:SAattenion}
\vspace{-16pt}
\end{figure}

\begin{table*}[!t]
    \centering
    \resizebox{0.85\textwidth}{!}{ 
    \begin{tabular}{c|c|c|c|c|c|c|c|c|c|c}
    \toprule[1pt]
    Datasets&Backbone& Method&$\#$Params (M)&$\#$FLOPs (G)& Abs Rel $\downarrow$& Sq Rel $\downarrow$&RMSE $\downarrow$ &$\delta_1 (\%)$ $\uparrow$ & $\delta_2 (\%)$ $\uparrow$ & $\delta_3 (\%)$ $\uparrow$\\
    \midrule[1pt]
    \multirow{18}*{M3D~\cite{Chang2017Matterport3DLF}}& \multirow{3}*{ResNet-18~\cite{He2015DeepRL}}&Equi-Baseline &14.39 &39.84 & 0.1428& 0.1203&0.5607&80.97 & 94.68 & 98.26 \\
    & &Ours&15.43 & 45.91 & 0.1272& 0.1070&0.5270 & 85.28& 95.28&98.49  \\
    & &\cellcolor{gray!40}$\varDelta$ &\cellcolor{gray!40}+1.04& \cellcolor{gray!40}+6.07& \cellcolor{gray!40}\textbf{-11.06$\%$}&\cellcolor{gray!40}\textbf{-9.73$\%$} &\cellcolor{gray!40}\textbf{-6.01$\%$} & \cellcolor{gray!40}\textbf{+4.31}& \cellcolor{gray!40}\textbf{+0.60}& \cellcolor{gray!40}\textbf{+0.23} \\
    \cmidrule{2-11}
    &\multirow{3}*{ResNet-34~\cite{He2015DeepRL}}&Equi-Baseline&24.50 &59.27 & 0.1255& 0.1048&0.5173&85.52 & 96.10 & 98.55 \\
     & &Ours&25.54 & 65.29 & 0.1115& 0.0914&0.4875 & 88.15& 96.46&98.74  \\
    & &\cellcolor{gray!40}$\varDelta$&\cellcolor{gray!40}+1.04& \cellcolor{gray!40}+6.02& \cellcolor{gray!40}\textbf{-11.16$\%$}&\cellcolor{gray!40}\textbf{-12.79$\%$} &\cellcolor{gray!40}\textbf{-5.76$\%$} & \cellcolor{gray!40}\textbf{+2.63}& \cellcolor{gray!40}\textbf{+0.36}& \cellcolor{gray!40}\textbf{+0.19} \\
    \cmidrule{2-11}
      &\multirow{3}*{ResNet-50$^*$~\cite{He2015DeepRL}} &Equi-Baseline& 29.63 & 75.13& 0.1370& 0.1209&0.5432& 82.51&94.83  & 98.07 \\
      & &Ours&42.99 &  170.11& 0.1112&0.0980 & 0.4870 &86.70 &96.01 & 98.61 \\
      &&\cellcolor{gray!40}$\varDelta$&\cellcolor{gray!40}+13.36& \cellcolor{gray!40}+94.98& \cellcolor{gray!40}\textbf{-18.75$\%$}&\cellcolor{gray!40}\textbf{-18.94$\%$} &\cellcolor{gray!40}\textbf{-10.35$\%$} & \cellcolor{gray!40}\textbf{+4.19}& \cellcolor{gray!40}\textbf{+1.18}& \cellcolor{gray!40}\textbf{+0.54} \\
      \cmidrule{2-11}
      & \multirow{3}*{EfficientNet-B5~\cite{Tan2019EfficientNetRM}}&Equi-Baseline&33.10& 18.81& 0.1034& 0.0831&0.4638& 89.92& 96.80& 98.80\\
      & &Ours&34.11 & 24.88& 0.1048& 0.0805 & 0.4524& 89.92&97.07 &99.14 \\
     & &\cellcolor{gray!40}$\varDelta$&\cellcolor{gray!40}+1.01& \cellcolor{gray!40}+6.07& \cellcolor{gray!40}\textbf{+1.35$\%$}&\cellcolor{gray!40}\textbf{-3.13$\%$} &\cellcolor{gray!40}\textbf{-2.46$\%$} & \cellcolor{gray!40}-0.00& \cellcolor{gray!40}\textbf{+0.27}& \cellcolor{gray!40}\textbf{+0.34} \\
      \cmidrule{2-11}
      & \multirow{3}*{SWin-B$^*$~\cite{Liu2021SwinTH}}&Equi-Baseline&  90.75&187.37 &0.1301 & 0.1250&0.5557 &84.06&94.76 &97.72 \\
      & &Ours&94.30 & 211.28 & 0.1249 & 0.1210  & 0.5462& 85.34 & 95.07& 97.77\\
      & &\cellcolor{gray!40}$\varDelta$& \cellcolor{gray!40}+3.55& \cellcolor{gray!40}+23.91& \cellcolor{gray!40}\textbf{-4.00$\%$}&\cellcolor{gray!40}\textbf{-3.20$\%$} &\cellcolor{gray!40}\textbf{-1.71$\%$} & \cellcolor{gray!40}\textbf{+1.28}& \cellcolor{gray!40}\textbf{+0.31}& \cellcolor{gray!40}\textbf{+0.05} \\
      \cmidrule{2-11}
      & \multirow{3}*{DilateFormer-T~\cite{jiao2023dilateformer}}&Equi-Baseline&19.93&46.96 &0.1429 &0.1274 &0.5748& 81.15&94.46 &97.79 \\
      & &Ours& 20.94 & 53.02& 0.1423&0.1251 &0.5517 &82.68 &95.00 &98.14 \\
      & &\cellcolor{gray!40}$\varDelta$& \cellcolor{gray!40}+1.01& \cellcolor{gray!40}+6.06& \cellcolor{gray!40}\textbf{-0.42$\%$}&\cellcolor{gray!40}\textbf{-1.81$\%$} &\cellcolor{gray!40}\textbf{-4.02$\%$} & \cellcolor{gray!40}\textbf{+1.53}& \cellcolor{gray!40}\textbf{+0.54}& \cellcolor{gray!40}\textbf{+0.35} \\
      \midrule[1pt]
      \multirow{6}*{S2D3D~\cite{Armeni2017Joint2D}}& \multirow{3}*{ResNet-34~\cite{He2015DeepRL}}&Equi-Baseline&24.50 & 59.22 & 0.1203& 0.0754 & 0.3724& 87.41 & 96.51& 98.77\\
      & &Ours&25.51 & 65.28 & 0.1182& 0.0728&0.3756 &88.72 &96.84 & 98.92\\
      & &\cellcolor{gray!40}$\varDelta$&\cellcolor{gray!40}+1.01& \cellcolor{gray!40}+6.06& \cellcolor{gray!40}\textbf{-9.21$\%$}&\cellcolor{gray!40}\textbf{-3.45$\%$} &\cellcolor{red!10}{\textbf{\textcolor{red}{+0.86$\%$}}} & \cellcolor{gray!40}\textbf{+1.31}& \cellcolor{gray!40}\textbf{+0.33}& \cellcolor{gray!40}\textbf{+0.15} \\
      \cmidrule{2-11}
      & \multirow{3}*{EfficientNet-B5~\cite{Tan2019EfficientNetRM}}&Equi-Baseline & 33.10& 18.81 & 0.1026&0.0638 &0.3580 & 89.43 & 97.06& 99.16\\
      & &Ours&34.11 & 24.88& 0.1018 &0.0603 & 0.3575& 89.47&97.23 &99.22 \\
      & &\cellcolor{gray!40}$\varDelta$& \cellcolor{gray!40}+1.01& \cellcolor{gray!40}+6.07& \cellcolor{gray!40}\textbf{-0.78$\%$}&\cellcolor{gray!40}\textbf{-5.49$\%$} &\cellcolor{gray!40}\textbf{-0.14$\%$} & \cellcolor{gray!40}\textbf{+0.03}& \cellcolor{gray!40}\textbf{+0.17}& \cellcolor{gray!40}\textbf{+0.06} \\
      \midrule[1pt]
      \multirow{6}*{Struct3D~\cite{Zheng2019Structured3DAL}}&\multirow{3}*{ResNet-34~\cite{He2015DeepRL}}&Equi-Baseline&24.50 &59.22& 0.2256&0.3910 &0.7641&76.90 &89.96 &94.49 \\
      & &Ours& 25.51&65.28&0.1480&0.2215&0.4961&87.41&94.34&96.66 \\
      & &\cellcolor{gray!40}$\varDelta$& \cellcolor{gray!40}+1.01& \cellcolor{gray!40}+6.06& \cellcolor{gray!40}\textbf{-34.40$\%$}&\cellcolor{gray!40}\textbf{-43.35$\%$} &\cellcolor{gray!40}\textbf{-35.07$\%$} & \cellcolor{gray!40}\textbf{+10.51}& \cellcolor{gray!40}\textbf{+4.38}& \cellcolor{gray!40}\textbf{+2.17} \\
      \cmidrule{2-11}
      & \multirow{3}*{EfficientNet-B5~\cite{Tan2019EfficientNetRM}}&Equi-Baseline&33.10&18.81 &0.1312 & 0.1938& 0.4312&88.53 &95.21 &97.40 \\
      & &Ours&34.11& 24.88 &0.1277 &0.1930 &0.4151 & 89.16&95.33 &97.43 \\
      &&\cellcolor{gray!40}$\varDelta$&\cellcolor{gray!40}+1.01& \cellcolor{gray!40}+6.07& \cellcolor{gray!40}\textbf{-2.67$\%$}&\cellcolor{gray!40}\textbf{-0.41$\%$} &\cellcolor{gray!40}\textbf{-3.73$\%$} & \cellcolor{gray!40}\textbf{+0.63}& \cellcolor{gray!40}\textbf{+0.12}& \cellcolor{gray!40}\textbf{+0.03} \\
      \bottomrule[1pt]
    \end{tabular}}
    \vspace{-5pt}
    \caption{\textbf{Quantitative comparison with ERP-based depth baseline}. For most conditions, we set the channel number $C$ to 64. Especially, we set $C$ to 256 for ResNet-50$^*$ and to 128 for Swin-B$^*$. \textbf{Bold} indicates performance improvement. \colorbox{red!10}{\textcolor{red}{\textbf{Red}}} indicates performance decline.}
    \vspace{-12pt}
    \label{tab:comparison-to-erp}
\end{table*}

\subsection{Bi-Projection Bi-Attention Fusion}
\label{sec:b2f}
In earlier methods~\cite{Wang2020BiFuseM3} and~\cite{Jiang2021UniFuseUF}, bi-projection feature fusion primarily depends on the geometric relationships between CP and ERP. They first apply the C2E function to re-project CP feature patches into ERP format feature maps and subsequently perform pixel-wise feature concatenation.
While this fusion is effective, it brings several problems: 1)
geometry-based re-projection and concatenation-based feature fusion significantly increase the computational costs (See Tab.~\ref{tab:comparison-to-soat}); 2) the geometry-based fusion process from CP patches to ERP panorama restricts ERP pixels from the global perception, where each ERP pixel perceives scene information only from its corresponding small-FoV CP patch without considering other patches; 3) this one-to-one alignment pattern only emphasizes spatial consistency, with no consideration for the semantic similarity. Thus, we design the Bi-Projection Bi-Attention Fusion (B2F) module to solve the above problems. As shown in Fig.~\ref{fig:overview}(b), B2F module first leverages semantic-aware affinity attention and distance-aware affinity attention to model the semantic and spatial dependencies between each ERP pixel-wise feature $F^{E}_{i,j}$ and ICOSAP point feature set $\boldsymbol{\{F^{I}_n\}_N}$. Here $(i,j)$ indicates the coordinate of a pixel in the ERP feature map, $i \in (1,h), j \in (1,w)$, and $N$ is the point number of the ICOSAP feature set. Consequently, a gated fusion block is employed to adaptively marry semantic- and distance-related information.
\begin{table*}[!t]
    \centering
    \resizebox{0.85\textwidth}{!}{ 
    \begin{tabular}{c|c|c|c|c|c|c|c|c|c|c|c}
    \toprule
    Datasets&Backbone& Method& Pub'Year &$\#$Params (M)&$\#$FLOPs (G)& Abs Rel $\downarrow$& Sq Rel $\downarrow$&RMSE $\downarrow$ &$\delta_1 (\%)$ $\uparrow$ & $\delta_2 (\%)$ $\uparrow$ & $\delta_3 (\%)$ $\uparrow$\\
    \midrule
    \multirow{16}*{M3D~\cite{Chang2017Matterport3DLF}}& \multirow{2}*{Transformer}&EGFormer~\cite{Yun2023EGformerEG} &ICCV'23 &15.39& 66.21& 0.1473&0.1517&0.6025&81.58&93.90&97.35\\
    &  &PanoFormer~\cite{Shen2022PanoFormerPT}&ECCV'22 & 20.38& 81.09& 0.1051&0.0966&0.4929&89.08&96.23&98.31\\
      \cmidrule{2-12}
     & \multirow{5}*{ResNet-18~\cite{He2015DeepRL}}& BiFuse~\cite{Wang2020BiFuseM3}& CVPR'20 &35.80&165.66& 0.1360& 0.1202& 0.5488&83.27 & 95.12 & 98.10 \\
     
     & &UniFuse~\cite{Jiang2021UniFuseUF}& RAL'21 &30.26&62.60 &\cellcolor{green!20} 0.1191& \cellcolor{green!20}0.1030& 0.5158&86.04 & \cellcolor{green!20}95.84 & 98.30 \\
     
     & &OmniFusion~\cite{Li2022OmniFusion3M}&CVPR'22 &32.35 & 98.68 & 0.1209 & 0.1090&\cellcolor{green!20}0.5055 & \cellcolor{green!20}86.58 &95.81 & 98.36 \\
     & &HRDFuse$^\dag$~\cite{Ai2023HRDFuseM3}&CVPR'23 &26.09 & 50.59 & 0.1414 & 0.1241&0.5507 & 81.48 &94.89 & 98.20 \\
     & &Ours&-&15.43 & 45.91 & 0.1272& 0.1070&0.5270 & 85.28& 95.28&\cellcolor{green!20}98.49  \\
     \cmidrule{2-12}
     &  \multirow{6}*{ResNet-34~\cite{He2015DeepRL}}& BiFuse~\cite{Wang2020BiFuseM3}&CVPR'20 &56.01&199.58& 0.1126& 0.0992& 0.5027&88.00 & 96.13 & 98.47 \\
    & & BiFuse++~\cite{Wang2022BiFuseSA}&TPAMI'22&52.49& 87.48 & 0.1123&  0.0915& 0.4853& 88.12 & 96.56 & 98.69 \\
     & & UniFuse~\cite{Jiang2021UniFuseUF}& RAL'21 &50.48&96.52 & 0.1144& 0.0936& \cellcolor{green!20}0.4835&87.85 & \cellcolor{green!20}96.59 & 98.73 \\
     
     & &OmniFusion~\cite{Li2022OmniFusion3M}&CVPR'22 &42.46 & 142.29 & 0.1161&0.1007&0.4931 & 87.72 &96.15 &98.44  \\
      & &HRDFuse$^\dag$~\cite{Ai2023HRDFuseM3}&CVPR'23 &46.31 & 80.87 & 0.1172 & 0.0971&0.5025 & 86.74 &96.17 & 98.49 \\
      & &Ours&-&25.54 & 65.29 & \cellcolor{green!20}0.1115& \cellcolor{green!20}0.0914&0.4875 & \cellcolor{green!20}88.15& 96.46&\cellcolor{green!20}98.74  \\
      \cmidrule{2-12}
      &\multirow{3}*{ResNet-50$^*$~\cite{He2015DeepRL}}& BiFuse~\cite{Wang2020BiFuseM3}& CVPR'20&253.08& 775.24& 0.1179&0.0981&0.4970 &86.74 & \cellcolor{green!20}96.27& \cellcolor{green!20}98.66\\
     & & UniFuse~\cite{Jiang2021UniFuseUF}&RAL'21&131.30& 222.30 & 0.1185&  0.0984& 0.5024& 86.66 & 96.18 & 98.50 \\
      & &Ours&-&42.99 &  170.11& \cellcolor{green!20}0.1112&\cellcolor{green!20}0.0980 & \cellcolor{green!20}0.4870 &\cellcolor{green!20}86.70 &96.01 & 98.61 \\
      \midrule[1pt]
  \multirow{6}*{S2D3D~\cite{Armeni2017Joint2D}}&  \multirow{2}*{Transformer}&EGFormer~\cite{Yun2023EGformerEG} & ICCV'23&15.39 & 66.21&0.1528 &0.1408 &0.4974& 81.85&93.38 &97.36 \\
      & &PanoFormer~\cite{Shen2022PanoFormerPT} &ECCV'22 & 20.38&81.09 &0.1122 & 0.0786&0.3945&88.74 &95.84 &98.59 \\
     \cmidrule{2-12}
     & \multirow{3}*{ResNet-34~\cite{He2015DeepRL}}& OmniFusion~\cite{Li2022OmniFusion3M}&CVPR'22 &42.46 &142.29& 0.1154&0.0775& 0.3809&86.74  &96.03& 98.71\\
     & & UniFuse~\cite{Jiang2021UniFuseUF}&RAL'21 &50.48 &96.52 &\cellcolor{green!20}0.1124&\cellcolor{green!20}0.0709 &\cellcolor{green!20}0.3555& 87.06 & \cellcolor{green!20}97.04& \cellcolor{green!20}98.99\\
      & &Ours&-&25.51 & 65.28 & 0.1182& 0.0728&0.3756 &\cellcolor{green!20}88.72 &96.84 & 98.92\\
      \midrule
      \multirow{5}*{Struct3D~\cite{Zheng2019Structured3DAL}}& \multirow{2}*{Transformer} &EGFormer~\cite{Yun2023EGformerEG} &ICCV'23&15.39 &66.21&0.2205 & 0.4509&0.6841 &79.79 & 90.71&94.55 \\
      & &PanoFormer~\cite{Shen2022PanoFormerPT} &ECCV'22&20.38& 81.09&0.2549&0.4949&0.7937 & 74.70 & 89.15&93.97 \\
     \cmidrule{2-12}
     &  \multirow{3}*{ResNet-34~\cite{He2015DeepRL}}&BiFuse~\cite{Wang2020BiFuseM3}&CVPR'20& 56.01&199.58&0.1573&0.2455 & 0.5213& 85.91& 94.00& 96.72\\
     & & UniFuse~\cite{Jiang2021UniFuseUF}&RAL'21& 50.48 &96.52 &0.1506 &0.2319 & 0.5016& 85.42& 93.99& \cellcolor{green!20}96.76 \\
      & &Ours&-& 25.51&65.28&\cellcolor{green!20}0.1480 &\cellcolor{green!20}0.2215&\cellcolor{green!20}0.4961&\cellcolor{green!20}87.41&\cellcolor{green!20}94.34&96.66 \\
      \bottomrule
    \end{tabular}}
    \vspace{-8pt}
    \caption{\textbf{Quantitative comparison with the SOTA methods}. $^\dag$ means that we modify the HRDFuse network structure for a fair comparison. \colorbox{green!30}{Green} represents the best performance under the given encoder backbone.
    }
    \vspace{-15pt}
    \label{tab:comparison-to-soat}
\end{table*}
\begin{figure}[!t]
\centering
\includegraphics[width=0.85\linewidth]{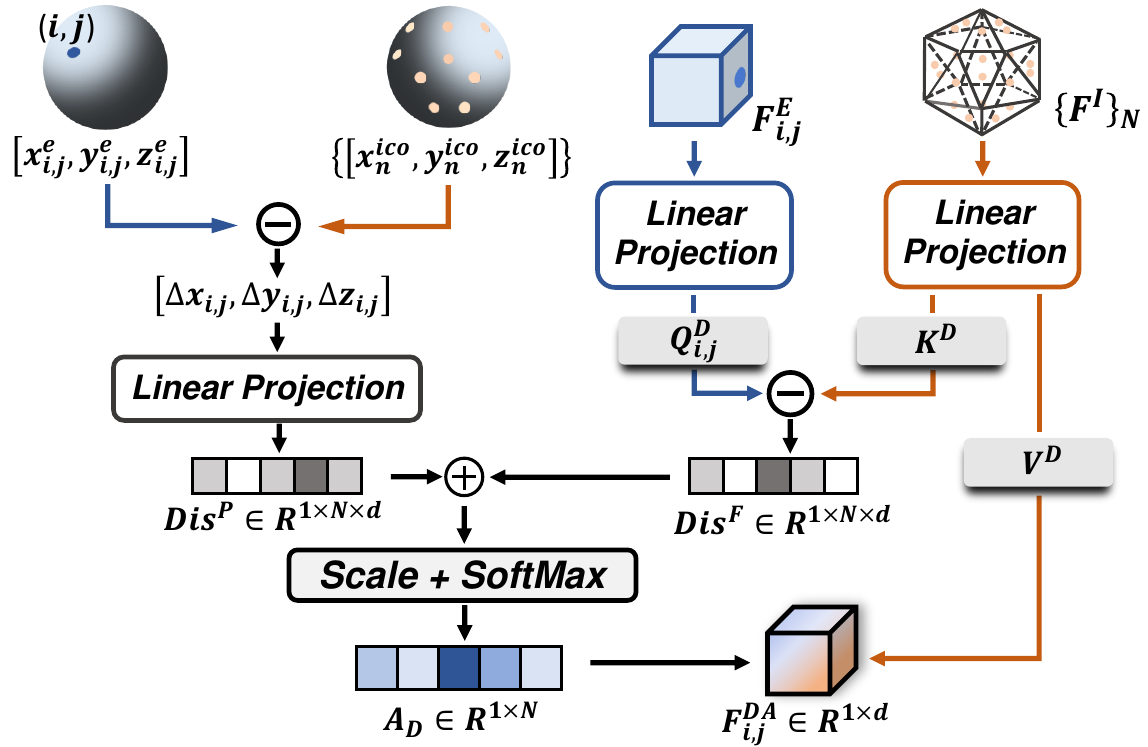}
\vspace{-3pt}
\caption{The architecture of distance-aware affinity attention, especially, $\mathbf{\textit{Q}}^{D}_{i,j} \in \mathbb{\textit{R}}^{1 \times d}$, $\mathbf{\textit{K}}^{D} \in \mathbb{R}^{N \times d}$, and $\mathbf{V}^{D} \in \mathbb{R}^{N \times d}$.}
\vspace{-14pt}
 \label{fig:DAattenion}
\end{figure}

\noindent \textbf{Semantic-aware affinity attention.} 
The semantic-aware affinity attention, as shown in Fig.~\ref{fig:SAattenion}, follows the the standard dot-product attention~\cite{dosovitskiy2021an}. In detail, given the extracted ERP feature map $F^{E} \in \mathbb{R}^{h\times w \times C}$ and ICOSAP point feature set $\boldsymbol{\{F^{I}_n\}_N} \in \mathbb{R}^{N \times C}$ (for simplicity, we denote it as $\boldsymbol{[F^{I}]}$), we generate the query $\mathbf{Q}^{S}_{i,j}$ from each ERP pixel-wise feature $F^{E}_{i,j}\in \mathbb{R}^{1 \times C}$ and produce the key $\mathbf{K}^{S}$ and value $\mathbf{V}^{S}$ from the whole ICOSAP point feature set $\boldsymbol{[F^{I}]}$. 
Then, we calculate the attention weight $\mathbf{A}^{S}_{i,j}$ based on $\mathbf{Q}^{S}_{i,j}$ and $\mathbf{K}^{S}$, and obtain the affinity feature $F^{SA}_{i,j}$ with $\ F^{SA}_{i,j} = \mathbf{A}^{S}_{i,j}*\mathbf{V}^{S}, \mathbf{A}^{S}_{i,j} = softmax(\frac{\mathbf{Q}^{S}_{i,j}{\mathbf{K}^{S}}^{T}}{\sqrt{d}}),$
where $d$ is the channel number and $d=C$. After querying all the ERP pixel-wise features, we can obtain the ERP format feature map $F^{SA}$ with the dimension of $\mathbb{R}^{h\times w\times d}$. The attention weight $A^{S}_{i,j}$ captures affinities based on the semantic-aware feature similarities, and the output $F^{SA}$ effectively integrates the global and local receptive fields.

\noindent \textbf{Distance-aware affinity attention.} 
Distance-aware affinity attention captures the differences between ERP and ICOSAP in both spatial and semantic information. This can better utilize the geometric prior knowledge of panoramas and precisely measure the distances between two different projection features. As depicted in Fig.~\ref{fig:DAattenion}, the distance-aware affinity attention is built upon the subtraction-based cross-attention mechanism~\cite{Wang2022DABERTDA}. Given the ERP pixel-wise feature $F^{E}_{i,j}$ and ICOSAP point feature set $\boldsymbol{[F^{I}]}$, we firstly calculate the spatial distance embedding $Dis^{SP}$ from the spatial coordinates of ERP pixel and ICOSAP point set, 
\ie $[x^{e}_{i,j},y^{e}_{i,j},z^{e}_{i,j}]$ and $\{[x^{ico}_{n},y^{ico}_{n},z^{ico}_{n}]\}_{N}$, as:
\vspace{-3pt}
\begin{equation}
    Dis^{SP}_{i,j} = [e^{-\varDelta x_{i,j}},e^{-\varDelta y_{i,j}},e^{-\varDelta z_{i,j}}]\mathbf{W}^{SP}, 
    \label{eq:pos}
\end{equation}
\vspace{-3pt}
where linear projection $\mathbf{W}^{SP}\in \mathbb{R}^{3\times d}$, $ Dis^{SP}_{i,j} \in \mathbb{R}^{1\times N\times d}$, and $[\varDelta x_{i,j},\varDelta y_{i,j},\varDelta z_{i,j}] \in \mathbb{R}^{1\times N\times 3}$ is the distances between $[x^{e}_{i,j},y^{e}_{i,j},z^{e}_{i,j}]$ and $\{[x^{ico}_{n},y^{ico}_{n},z^{ico}_{n}]\}_{N}$. In particular, the operation $e^{-(\cdot)}$ allows $Dis^{SP}$ to pay more attention to the close parts between ERP pixels and ICOSAP point set. After that, we produce the query $\mathbf{Q}^{D}_{i,j}$ and key $\mathbf{K}^{D}$ from $F^{E}_{i,j}$ and $\boldsymbol{[F^{I}]}$, respectively and calculate the semantic distance embedding $Dis^{SE}_{i,j}$:
\begin{equation}
\setlength\abovedisplayskip{3pt}
\setlength\belowdisplayskip{3pt}
 Dis^{SE}_{i,j} = e^{-\left \| \mathbf{Q}^{D}_{i,j}- \mathbf{K}^{D}\right \|}, \label{eq:dis}
\end{equation}
where $\mathbf{W}^{D}_{Q}$, $\mathbf{W}^{D}_{K}\in \mathbb{R}^{C \times d}$ are linear projections, 
$\mathbf{Q}^{D}_{i,j} \in \mathbb{R}^{1 \times d}$, $\mathbf{K}^{D} \in \mathbb{R}^{N \times d}$, and $Dis^{SE}_{i,j} \in \mathbb{R}^{1\times N \times d}$. Lastly, the distance-aware attention weight $\mathbf{A}^{D}_{i,j}$ is generated with spatial and semantic distance embeddings, and the distance-aware affinity feature vector $F_{i,j}^{DA}$ is obtained from the attention weight $\mathbf{A}^{D}_{i,j}$ 
and the value $\mathbf{V}^{D}$:
\begin{equation}
\footnotesize
\setlength\abovedisplayskip{3pt}
\setlength\belowdisplayskip{3pt}
    \mathbf{A}^{D}_{i,j} = softmax(\frac{ \small{\sum} (Dis^{SP}_{i,j} + Dis^{SE}_{i,j})}{\sqrt{d}}),
\end{equation}
\begin{equation}
\footnotesize
\setlength\abovedisplayskip{3pt}
\setlength\belowdisplayskip{3pt}
     \mathbf{V}^{D} = F^{I}\mathbf{W}^{D}_{V}, \ \
     F^{DA}_{i,j} = \mathbf{A}^{D}_{i,j}*\mathbf{V}^{D},
\end{equation}
where $\small{\sum} $ means the sum for the channel dimension, \ie $\small{\sum(Dis^{SP}_{i,j} + Dis^{SE}_{i,j}) \in \mathbb{R}^{1\times N}}$. After querying all ERP pixel-wise features, we obtain the distance-aware aggregated feature $F^{DA}\in\mathbb{R}^{h\times w \times d}$, $d=C$.

\noindent \textbf{Gated fusion.} Since direct average or concatenation may compromise the original representation ability, inspired by~\cite{Cheng2017LocalitySensitiveDN}, we propose the gated fusion block to adaptively fuse $F^{SA}$ and $F^{DA}$ and obtain the representations $F^{GL}$ from a local-with-global perspective, formulated as:
\vspace{-4pt}
\begin{small}
\begin{gather}
    F^{GL}= g^{SA} * F^{SA} + g^{DA} * F^{DA},\label{eq:gate_fusion}\\
    g^{SA} = \sigma_{SA} (W^{SA}_g\cdot[F^{SA};F^{DA}]),\nonumber \\
    g^{DA} = \sigma_{DA} (W^{DA}_g\cdot[F^{SA};F^{DA}]),\nonumber
\end{gather}
\end{small}
where $W^{SA}_g$ and $W^{DA}_g$ are linear projections, $\sigma_{SA}(\cdot)$ and $\sigma_{DA}(\cdot)$ are the sigmoid functions.

\begin{figure*}[t!]
    \centering
    \includegraphics[width=0.89\textwidth]{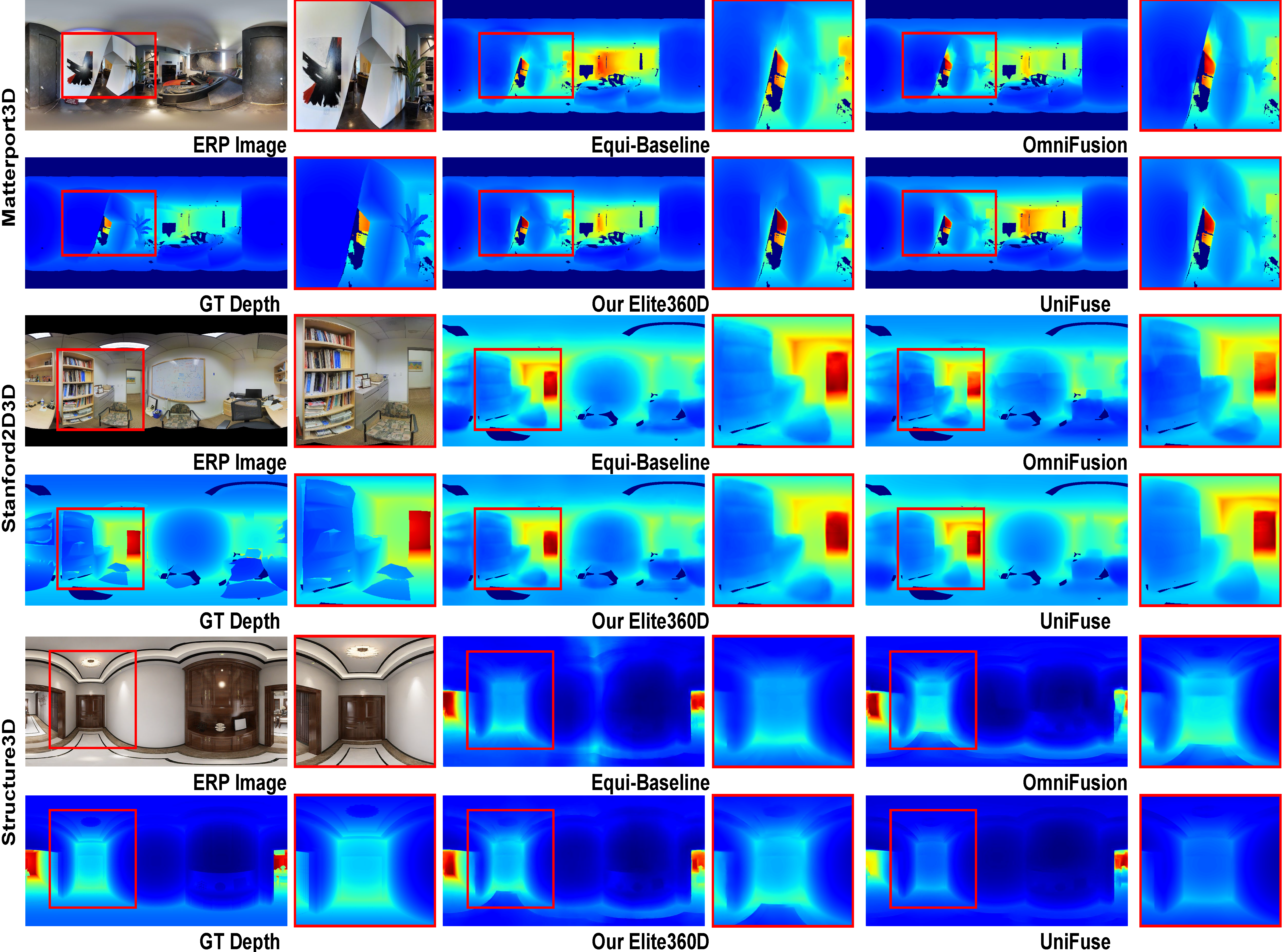}
     \vspace{-8pt}
\caption{Qualitative results (with ResNet-34 as the backbone) on Matterport3D (top), Stanford2D3D (middle) and Structure3D (bottom).}
    \label{fig:compare}
     \vspace{-12pt}
\end{figure*}
\subsection{Optimization}
\label{sec:loss}

With the fused feature $F^{GL}$ and multi-scale ERP feature maps in the ERP encoder backbone, we feed them into a decoder~\cite{Ronneberger2015UNetCN} with several up-sampling blocks and skip-connections to output the final depth. For the depth supervision, we follow existing works~\cite{Shen2022PanoFormerPT, Pintore2021SliceNetDD} and employ the combination of Berhu loss~\cite{Laina2016DeeperDP} and gradient loss~\cite{Pintore2021SliceNetDD}.
(\textit{Details of training loss can be found in the suppl. material}.).

\vspace{-5pt}
\section{Experiments}

\vspace{-3pt}
\subsection{Datasets, Metrics, and Implementation Details}
\vspace{-3pt}
\noindent\textbf{Datasets and Metrics.} We evaluate Elite360D on three datasets: two real-world datasets, Matterport3D(M3D)~\cite{Chang2017Matterport3DLF}, Stanford2D3D(S2D3D)~\cite{Armeni2017Joint2D}, and a recently proposed large-scale synthetic dataset, Structured3D(Struct3D)~\cite{Zheng2019Structured3DAL}. For the evaluation metrics, we follow previous works~\cite{Jiang2021UniFuseUF,Shen2022PanoFormerPT,Yun2023EGformerEG} to employ some standard metrics, including absolute relative error (Abs Rel), squared relative error (Sq Rel), root mean squared error (RMSE), and three threshold percentage $\delta<\alpha^{t}$ $(\alpha=1.25, t=1,2,3)$, denoted as $\delta_{t}$. Additionally, we measure the number of parameters and FLOPS to evaluate the efficiency of our method.
\begin{table}[!t]
    \centering
    \resizebox{0.82\linewidth}{!}{ 
    \begin{tabular}{c|c|cccc}
    \toprule
    Backbone & Pre-trained&Abs Rel$\downarrow$&Sq Rel$\downarrow$&RMSE$\downarrow$& $\delta_1$ $\uparrow$ \\
    \midrule
    \multirow{2}*{ResNet-34}& \XSolidBrush  & 0.1596 &0.1452& 0.5856 &81.36 \\
    & \Checkmark  & \cellcolor{gray!20}0.1115 &\cellcolor{gray!20}0.0914&\cellcolor{gray!20}0.4875 &\cellcolor{gray!20}88.15 \\
    \midrule
    \multirow{2}*{EfficientNet-B5} & \XSolidBrush & 0.1211 &0.1087 &0.5131 &86.80 \\
    & \Checkmark & \cellcolor{gray!20}0.1048 &\cellcolor{gray!20}0.0805& \cellcolor{gray!20}0.4524&\cellcolor{gray!20}89.92\\
    \midrule
    \multirow{2}*{DilateFormer} &\XSolidBrush  & 0.1515 &0.1415 &0.5694&80.95 \\
    & \Checkmark & \cellcolor{gray!20}0.1423 &\cellcolor{gray!20}0.1251& \cellcolor{gray!20}0.5517&\cellcolor{gray!20}82.68 \\
    \bottomrule
    \end{tabular}}
    \vspace{-8pt}
    \caption{The impact of pre-training on ERP encoder backbones.}
    \label{tab:ab_erp_pretrain}
    \vspace{-20pt}
\end{table}

\noindent\textbf{Training Details.}
We use diverse ERP encoder backbones, including CNNs (ResNet-18, 34, 50~\cite{He2015DeepRL}, EfficientNet B5~\cite{Tan2019EfficientNetRM}), and transformers (Swin-B~\cite{Liu2021SwinTH}, DilateFormer-T~\cite{jiao2023dilateformer}). All backbones are pre-trained on ImageNet-1K~\cite{JiaDeng2009ImageNetAL}. We set the default channel number $C$ to 64 and default subdivision level of ICOSAP as $l=4$. For the ICOSAP encoder, we employ the one of Point transformer~\cite{Zhao2020PointT} with three down-sample blocks. Following~\cite{Jiang2021UniFuseUF}, we use Adam optimizer~\cite{Kingma2014AdamAM} and a constant learning rate of 1e$^{-4}$. Considering the unfair comparisons stemming from variations in hyper-parameters and validation procedures used across different methods, we re-train the existing methods from scratch and validate them, following the unified training and validation settings~\cite{Jiang2021UniFuseUF}. (\textit{Due to page limit, detailed training and validation settings can be found in suppl. mat.}). 
\vspace{-6pt}
\subsection{Performance Comparison}
\vspace{-3pt}

\noindent \textbf{Comparisons with ERP-based depth baselines.} As shown in Tab.~\ref{tab:comparison-to-erp}, with an increase of only $\thicksim$1M parameters ($C$=64), our Elite360D demonstrates substantial advancements over the ERP-based baselines across different ERP encoder backbones on all three datasets. Specifically, for the Matterport3D dataset, Elite360D achieves reductions exceeding 10$\%$ in Abs Rel error (ResNet-18, 34), along with reductions of 4.00$\%$ in Abs Rel error (Swin-B) and 4.02$\%$ in RMSE error (DilateFormer-T). Besides, with the larger channel number $C=256$ (ResNet-50), Elite360D outperforms ERP baseline by 18.75$\%$ (Abl Rel), 18.94$\%$ (Sq Rel). For the small-scale S2D3D dataset, Elite360D outperforms ERP baseline by 9.21$\%$ in Abs Rel error and 1.31$\%$ in accuracy $\delta_1$ (ResNet-34), as well as 5.49$\%$ in Sq Rel error (EfficientNet-B5). Remarkably, on the larger-scale Structure3D, Elite360 performs favorably against the baseline by a significant margin, especially with ResNet-34. 

\noindent \textbf{Comparisons with prevalent methods.} In Tab.~\ref{tab:comparison-to-soat}, we conduct a comprehensive comparison with prevalent supervised methods. From the results, we can observe that our approach achieves similar or even superior performance compared to existing both bi-projection fusion methods and single input methods at a significantly lower cost, particularly on two large-scale datasets, Matterport3D and Structure3D. Specifically, for the Matterport3D dataset, our Elite360D with ResNet-34 outperforms UniFuse by 2.53$\%$ (Abs Rel) and with ResNet-50 outperforms BiFuse by 2.01$\%$ (RMSE). For the Structure3D dataset, our Elite360D with ResNet-34 outperforms UniFuse by 4.48$\%$ (Seq Rel), 1.99$\%$ ($\delta_1$). For performance on the Stanford2D3D dataset, we suspect it might be related to the ICOSAP point encoder. The limited data of Stanford2D3D dataset restricts the ability of the transformer-based point encoder to provide accurate global perception. Moreover, in Fig.~\ref{fig:compare}, we present the qualitative comparisons. Our Elite360D can predict more accurate depth values based on the local-with-global perception capabilities (e.g., flowers, shelves and doors). \textit{Additional qualitative results and inference time comparisons can be found in the suppl. material.}

\vspace{-3pt}
\subsection{Ablation Study and Analyses}
\vspace{-3pt}
Most of ablation experiments are conducted on the Matteport3D test dataset with ResNet34 as the backbone .
\begin{table}[!t]
    \centering
    \resizebox{0.8\linewidth}{!}{ 
    \begin{tabular}{c|c|c|c|c}
    \toprule
    Bi-projection feature fusion & Abs Rel $\downarrow$&Sq Rel $\downarrow$&RMSE $\downarrow$& $\delta_1$ $\uparrow$\\
    \midrule
    SFA~\cite{Ai2023HRDFuseM3} + Add & 0.1276& 0.1002 &0.5150 & 84.27\\
    SFA~\cite{Ai2023HRDFuseM3} + Concat & 0.1191& 0.1019 & 0.5143& 86.52\\
    Only SA & 0.1204& 0.1014 &  0.5121& 86.26\\
    Only DA &  0.1184 & 0.0972 & 0.4944 &87.06 \\
    Our B2F (SA + DA) & \textbf{0.1115} & \textbf{0.0914} & \textbf{0.4875} &  \textbf{88.15} \\
    \bottomrule
    \end{tabular}}
    \vspace{-8pt}
    \caption{The ablation results for B2F module.}
    \label{tab:ab_biprojection_fusion}
    \vspace{-5pt}
\end{table}
\begin{table}[!t]
    \centering
    \resizebox{0.8\linewidth}{!}{ 
    \begin{tabular}{c|c|c|c|c}
    \toprule
    Final fusion & Abs Rel $\downarrow$&Sq Rel $\downarrow$&RMSE $\downarrow$& $\delta_1$ $\uparrow$\\
    \midrule
    Add &0.1685 & 0.1481 & 0.5809& 74.60\\
    Average & 0.1198 & 0.0918 & 0.4893&86.65 \\
    Concatenation &0.1145 & 0.0937& 0.4880&87.66 \\
    Adaptive fusion~\cite{Ai2023HRDFuseM3} & 0.1244&   0.0968& 0.4891& 86.08\\
    Our gated fusion& \textbf{0.1115} & \textbf{0.0914} & \textbf{0.4875} &  \textbf{88.15} \\
    \bottomrule
    \end{tabular}}
    \vspace{-8pt}
    \caption{The ablation results for the fusion of B2F module.}
    \label{tab:ab_final_fusion}
    \vspace{-15pt}
\end{table}

\noindent \textbf{The Effect of pre-training.}
We verify the effectiveness of ImageNet~\cite{JiaDeng2009ImageNetAL} pre-training with different encoder backbones. As observed from Tab.~\ref{tab:ab_erp_pretrain}, the pre-training results in a significant improvement for all encoder backbones, \eg, 
6.79$\%$ improvement in accuracy $\delta_1$ (ResNet-34). Notably, pre-training has a relatively small impact on DilateFormer. Combined with the results in Tab.~\ref{tab:comparison-to-erp}, the explanation of this phenomenon is that the default input resolution in pre-trained models is different from actual input, thereby impacting the resolution-related position embeddings. In general, pre-training based on large-scale perspective images can effectively enhance the performance of models based on 360$^\circ$ images and reduce the risk of overfitting.

\noindent \textbf{The effectiveness of B2F module.} In Tab.~\ref{tab:ab_biprojection_fusion}, we compare four available bi-projection feature fusion modules. To align the spatial dimensions between ICOSAP point feature set and ERP feature map, we introduce SFA module from~\cite{Ai2023HRDFuseM3}. After that, we employ direct addition and concatenation to aggregate these two projections. We also achieve the bi-projection feature fusion with semantic-aware affinity attention (SA) alone and distance-aware affinity attention (DA) alone. Compared to the methods based solely on semantic-aware feature similarities (The first three rows), single distance-aware affinity attention can achieve better performance, which indicates that the spatial positional relationships boost the bi-projection feature fusion. Overall, our B2F module achieves the best performance.

\noindent \textbf{The superiority of ICOSAP.} As only CP$/$TP's patch centers lie on the sphere's surface, we extract the feature embedding from each CP$/$TP patch and employs the patch center coordinates and feature embedding as the input of B2F module. In Tab.~\ref{tab:ab_diff_proj}, we show the results with ResNet18 backbone. Our Elite360D, utilizing the ICOSAP point set, marginally outperforms models with CP and TP patches, while exhibiting fewer parameters and FLOPs.

\noindent \textbf{The effectiveness of gated fusion.} We conduct an ablation study for the gated fusion block, outlined in Tab.~\ref{tab:ab_final_fusion}. With the feature maps $F^{SA}$ and $F^{DA}$, We compare it with the direct addition, average fusion, concatenation, and the adaptive fusion in~\cite{Ai2023HRDFuseM3}. The gated fusion performs best.

\begin{table}[!t]
\centering
\resizebox{1\linewidth}{!}{ 
    \begin{tabular}{cccccc}
    \toprule
    Method&$\#$Param(M) &$\#$FLOPs(G) &Abs Rel $\downarrow$ & RMSE $\downarrow$  &$\delta_1$ $\uparrow$ \\
    \cmidrule{1-6}
   ERP-CP& 25.66 &  54.15 &  0.1369& 0.5401& 83.69\\
   \cmidrule{1-6}
   ERP-TP (N=18) & 25.66& 50.58 & 0.1328& 0.5385& 83.87 \\
    \cmidrule{1-6}
    ERP-ICOSAP (Ours) &\textbf{15.43}&\textbf{45.91} &\textbf{0.1272} & \textbf{0.5270} &\textbf{85.28}\\
    \bottomrule
    \end{tabular}}
    \vspace{-5pt}
    \caption{The comparison of different projections on Matterport3D.}
    \label{tab:ab_diff_proj}
    \vspace{-8pt}
\end{table}
\begin{table}[!t]
    \centering
    \resizebox{\linewidth}{!}{ 
    \begin{tabular}{c|c|c|c|c|c|c}
    \toprule
    $N$ of $\{F^{I}\}$ & $\#$Params (M)& $\#$FLOPs (G)&Abs Rel$\downarrow$&Sq Rel$\downarrow$&RMSE$\downarrow$& $\delta_1$$\uparrow$ \\
    \midrule
    20 & 27.41& 66.29& 0.1157 & 0.0995 &0.5024 &87.12 \\
    80 & 25.54& 65.29 & \textbf{0.1115} & \textbf{0.0914} & \textbf{0.4875} &  \textbf{88.15} \\
    320 &24.98 & 64.32 & 0.1153 & 0.0943 & 0.4905 & 87.85 \\
    \bottomrule  
    \end{tabular}}
    \vspace{-8pt}
    \caption{Impact of the ICOSAP point-wise feature number $N$. Larger $N$, fewer down-sampling blocks in the point encoder.}
    \label{tab:ICOSAP_point}
    \vspace{-15pt}
\end{table}

\noindent \textbf{ICOSAP point feature number $N$.} We study the effect of the ICOSAP feature point number (See Tab.~\ref{tab:ICOSAP_point}). Too few points (N=20) lead to the over-concentrated global contextual information resulting from excessive down-sampling blocks, while too many points (N=320) lead to under-concentrated condition, resulting in insufficient perception of ERP pixel features. Best performance can be observed when N=80 and we used N=80 as default in this paper.
\vspace{-5pt}
\section{Conclusion and Future Work}
\vspace{-5pt}
In this paper, we proposed a novel bi-projection fusion solution for efficient 360 depth estimation. To address the limited local receptive field of ERP pixel-wise features and avoid expensive bi-projection fusion modules, 
we proposed a compact yet effective B2F module to learn the representations with local-with-global perceptions from ERP and ICOSAP.
With an increase of 1M parameters, we significantly improved the performance of the ERP-based depth estimation baseline. Remarkably, our approach achieved performance on par with complex state-of-the-art methods.
\noindent \textbf{Future Work:} From the experimental results, we observed that ERP-based depth baseline, with pre-trained EfficientNet backbone, even outperforms existing specifically designed methods. Therefore, in the future, we will explore how to fully leverage different projections and successful perspective models for 360$^\circ$ community.

\vspace{-4pt}
\section*{Acknowledgement}
\vspace{-4pt}
This paper is supported by the National Natural Science Foundation of China (NSF) under Grant No. NSFC22FYT45 and the Guangzhou City, University and Enterprise Joint Fund under Grant No.SL2022A03J01278.

\clearpage
{
    \small
    \bibliographystyle{ieeenat_fullname}
    \bibliography{main}

\begin{thebibliography}{51}
\providecommand{\natexlab}[1]{#1}
\providecommand{\url}[1]{\texttt{#1}}
\expandafter\ifx\csname urlstyle\endcsname\relax
  \providecommand{\doi}[1]{doi: #1}\else
  \providecommand{\doi}{doi: \begingroup \urlstyle{rm}\Url}\fi

\bibitem[Ai et~al.(2022)Ai, Cao, Zhu, Bai, Chen, and Wang]{2022DeepLFAi}
Hao Ai, Zidong Cao, Jin Zhu, Haotian Bai, Yucheng Chen, and Ling Wang.
\newblock Deep learning for omnidirectional vision: A survey and new perspectives.
\newblock \emph{ArXiv}, abs/2205.10468, 2022.

\bibitem[Ai et~al.(2023)Ai, Cao, Cao, Shan, and Wang]{Ai2023HRDFuseM3}
Hao Ai, Zidong Cao, Yan-Pei Cao, Ying Shan, and Lin Wang.
\newblock Hrdfuse: Monocular 360° depth estimation by collaboratively learning holistic-with-regional depth distributions.
\newblock \emph{2023 IEEE/CVF Conference on Computer Vision and Pattern Recognition (CVPR)}, pages 13273--13282, 2023.

\bibitem[Armeni et~al.(2017)Armeni, Sax, Zamir, and Savarese]{Armeni2017Joint2D}
Iro Armeni, Sasha Sax, Amir~Roshan Zamir, and Silvio Savarese.
\newblock Joint 2d-3d-semantic data for indoor scene understanding.
\newblock \emph{CoRR}, abs/1702.01105, 2017.

\bibitem[Cao et~al.(2023)Cao, Ai, Cao, Shan, Qie, and Wang]{Cao2023OmniZoomerLT}
Zidong Cao, Hao Ai, Yan Cao, Ying Shan, Xiaohu Qie, and Lin Wang.
\newblock Omnizoomer: Learning to move and zoom in on sphere at high-resolution.
\newblock \emph{2023 IEEE/CVF International Conference on Computer Vision (ICCV)}, pages 12851--12861, 2023.

\bibitem[Chang et~al.(2017)Chang, Dai, Funkhouser, Halber, Nie{\ss}ner, Savva, Song, Zeng, and Zhang]{Chang2017Matterport3DLF}
Angel~X. Chang, Angela Dai, Thomas~A. Funkhouser, Maciej Halber, Matthias Nie{\ss}ner, Manolis Savva, Shuran Song, Andy Zeng, and Yinda Zhang.
\newblock Matterport3d: Learning from {RGB-D} data in indoor environments.
\newblock In \emph{3DV}, pages 667--676. {IEEE} Computer Society, 2017.

\bibitem[Chen et~al.(2022)Chen, Li, Zhang, Fang, Jiang, Zhao, Zhou, and Zhao]{Chen2022AutoAlignPF}
Zehui Chen, Zhenyu Li, Shiquan Zhang, Liangji Fang, Qinghong Jiang, Feng Zhao, Bolei Zhou, and Hang Zhao.
\newblock Autoalign: Pixel-instance feature aggregation for multi-modal 3d object detection.
\newblock In \emph{International Joint Conference on Artificial Intelligence}, 2022.

\bibitem[Cheng et~al.(2018)Cheng, Chao, Dong, Wen, Liu, and Sun]{Cheng2018CubePF}
Hsien-Tzu Cheng, Chun-Hung Chao, Jin-Dong Dong, Hao-Kai Wen, Tyng-Luh Liu, and Min Sun.
\newblock Cube padding for weakly-supervised saliency prediction in 360° videos.
\newblock \emph{2018 IEEE/CVF Conference on Computer Vision and Pattern Recognition}, pages 1420--1429, 2018.

\bibitem[Cheng et~al.(2020)Cheng, Wang, Zhou, Guan, and Yang]{Cheng2020ODECNNOD}
Xinjing Cheng, Peng Wang, Yanqi Zhou, Chenye Guan, and Ruigang Yang.
\newblock Omnidirectional depth extension networks.
\newblock In \emph{2020 IEEE International Conference on Robotics and Automation (ICRA)}, pages 589--595, 2020.

\bibitem[Cheng et~al.(2017)Cheng, Cai, Li, Zhao, and Huang]{Cheng2017LocalitySensitiveDN}
Yanhua Cheng, Rui Cai, Zhiwei Li, Xin Zhao, and Kaiqi Huang.
\newblock Locality-sensitive deconvolution networks with gated fusion for rgb-d indoor semantic segmentation.
\newblock \emph{2017 IEEE Conference on Computer Vision and Pattern Recognition (CVPR)}, pages 1475--1483, 2017.

\bibitem[Cohen et~al.(2019)Cohen, Weiler, Kicanaoglu, and Welling]{Cohen2019GaugeEC}
Taco Cohen, Maurice Weiler, Berkay Kicanaoglu, and Max Welling.
\newblock Gauge equivariant convolutional networks and the icosahedral cnn.
\newblock In \emph{International Conference on Machine Learning}, 2019.

\bibitem[Deng et~al.(2009)Deng, Dong, Socher, Li, Li, and Fei-Fei]{JiaDeng2009ImageNetAL}
Jia Deng, Wei Dong, Richard Socher, Li-Jia Li, Kai Li, and Li Fei-Fei.
\newblock Imagenet: A large-scale hierarchical image database.
\newblock \emph{computer vision and pattern recognition}, 2009.

\bibitem[Dosovitskiy et~al.(2021)Dosovitskiy, Beyer, Kolesnikov, Weissenborn, Zhai, Unterthiner, Dehghani, Minderer, Heigold, Gelly, Uszkoreit, and Houlsby]{dosovitskiy2021an}
Alexey Dosovitskiy, Lucas Beyer, Alexander Kolesnikov, Dirk Weissenborn, Xiaohua Zhai, Thomas Unterthiner, Mostafa Dehghani, Matthias Minderer, Georg Heigold, Sylvain Gelly, Jakob Uszkoreit, and Neil Houlsby.
\newblock An image is worth 16x16 words: Transformers for image recognition at scale.
\newblock In \emph{International Conference on Learning Representations}, 2021.

\bibitem[Eder et~al.(2019)Eder, Shvets, Lim, and Frahm]{Eder2019TangentIF}
Marc Eder, Mykhailo Shvets, John Lim, and Jan-Michael Frahm.
\newblock Tangent images for mitigating spherical distortion.
\newblock \emph{2020 IEEE/CVF Conference on Computer Vision and Pattern Recognition (CVPR)}, pages 12423--12431, 2019.

\bibitem[He et~al.(2015)He, Zhang, Ren, and Sun]{He2015DeepRL}
Kaiming He, X. Zhang, Shaoqing Ren, and Jian Sun.
\newblock Deep residual learning for image recognition.
\newblock \emph{2016 IEEE Conference on Computer Vision and Pattern Recognition (CVPR)}, pages 770--778, 2015.

\bibitem[Jiang et~al.(2019)Jiang, Huang, Kashinath, Prabhat, Marcus, and Niessner]{Jiang2019SCUG}
ChiyuMax Jiang, Jingwei Huang, Karthik Kashinath, Prabhat Prabhat, Philip Marcus, and Matthias Niessner.
\newblock Spherical cnns on unstructured grids.
\newblock \emph{International Conference on Learning Representations,International Conference on Learning Representations}, 2019.

\bibitem[Jiang et~al.(2021)Jiang, Sheng, Zhu, Dong, and Huang]{Jiang2021UniFuseUF}
Hualie Jiang, Zhe Sheng, Siyu Zhu, Zilong Dong, and Rui Huang.
\newblock Unifuse: Unidirectional fusion for 360° panorama depth estimation.
\newblock \emph{IEEE Robotics and Automation Letters}, 6:\penalty0 1519--1526, 2021.

\bibitem[Jiao et~al.(2023)Jiao, Tang, Lin, Gao, Ma, Wang, and Zheng]{jiao2023dilateformer}
Jiayu Jiao, Yu-Ming Tang, Kun-Yu Lin, Yipeng Gao, Jinhua Ma, Yaowei Wang, and Wei-Shi Zheng.
\newblock Dilateformer: Multi-scale dilated transformer for visual recognition.
\newblock \emph{IEEE Transactions on Multimedia}, pages 1--14, 2023.

\bibitem[Kingma and Ba(2014)]{Kingma2014AdamAM}
Diederik~P. Kingma and Jimmy Ba.
\newblock Adam: A method for stochastic optimization.
\newblock \emph{CoRR}, abs/1412.6980, 2014.

\bibitem[Kumar et~al.(2021)Kumar, Yogamani, Rashed, Sitsu, Witt, Leang, Milz, and M{\"a}der]{RaviKumar2021OmniDetSV}
Varun~Ravi Kumar, Senthil~Kumar Yogamani, Hazem Rashed, Ganesh Sitsu, Christian Witt, Isabelle Leang, Stefan Milz, and Patrick M{\"a}der.
\newblock Omnidet: Surround view cameras based multi-task visual perception network for autonomous driving.
\newblock \emph{IEEE Robotics and Automation Letters}, 6:\penalty0 2830--2837, 2021.

\bibitem[Laina et~al.(2016)Laina, Rupprecht, Belagiannis, Tombari, and Navab]{Laina2016DeeperDP}
Iro Laina, C. Rupprecht, Vasileios Belagiannis, Federico Tombari, and Nassir Navab.
\newblock Deeper depth prediction with fully convolutional residual networks.
\newblock \emph{3DV 2016}, pages 239--248, 2016.

\bibitem[Lee et~al.(2018)Lee, Jeong, Yun, Cho, and jin Yoon]{Lee2018SpherePHDAC}
Yeonkun Lee, Jaeseok Jeong, Jong~Seob Yun, Wonjune Cho, and Kuk jin Yoon.
\newblock Spherephd: Applying cnns on a spherical polyhedron representation of 360° images.
\newblock \emph{2019 IEEE/CVF Conference on Computer Vision and Pattern Recognition (CVPR)}, pages 9173--9181, 2018.

\bibitem[Li et~al.(2023)Li, Wang, Yuan, Shen, Sheng, and Dong]{Li2023mathcalA}
Meng Li, Senbo Wang, Weihao Yuan, Weichao Shen, Zhe Sheng, and Zilong Dong.
\newblock $\mathcal{S}^{2}$net: Accurate panorama depth estimation on spherical surface.
\newblock \emph{IEEE Robotics and Automation Letters}, 8:\penalty0 1053--1060, 2023.

\bibitem[Li et~al.(2021)Li, Shen, Gao, Zhu, Zhai, and Guo]{Li2021LookingHO}
Yunhao Li, Wei Shen, Zhongpai Gao, Yucheng Zhu, Guangtao Zhai, and Guodong Guo.
\newblock Looking here or there? gaze following in 360-degree images.
\newblock \emph{2021 IEEE/CVF International Conference on Computer Vision (ICCV)}, pages 3722--3731, 2021.

\bibitem[Liu et~al.(2021)Liu, Lin, Cao, Hu, Wei, Zhang, Lin, and Guo]{Liu2021SwinTH}
Ze Liu, Yutong Lin, Yue Cao, Han Hu, Yixuan Wei, Zheng Zhang, Stephen Lin, and Baining Guo.
\newblock Swin transformer: Hierarchical vision transformer using shifted windows.
\newblock \emph{2021 IEEE/CVF International Conference on Computer Vision (ICCV)}, pages 9992--10002, 2021.

\bibitem[Man et~al.(2023)Man, Gui, and Wang]{Man2023BEVGuidedMF}
Yunze Man, Liangyan Gui, and Yu-Xiong Wang.
\newblock Bev-guided multi-modality fusion for driving perception.
\newblock \emph{2023 IEEE/CVF Conference on Computer Vision and Pattern Recognition (CVPR)}, pages 21960--21969, 2023.

\bibitem[Martin et~al.(2021)Martin, Serrano, Bergman, Wetzstein, and Masi{\'a}]{Martin2021ScanGAN360AG}
Daniel Martin, Ana Serrano, Alexander~W. Bergman, Gordon Wetzstein, and Bel{\'e}n Masi{\'a}.
\newblock Scangan360: A generative model of realistic scanpaths for 360° images.
\newblock \emph{IEEE Transactions on Visualization and Computer Graphics}, 28:\penalty0 2003--2013, 2021.

\bibitem[Monroy et~al.(2017)Monroy, Lutz, Chalasani, and Smolic]{Monroy2017SalNet360SM}
Rafael Monroy, Sebastian Lutz, Tejo Chalasani, and Aljoscha Smolic.
\newblock Salnet360: Saliency maps for omni-directional images with cnn.
\newblock \emph{Signal Process. Image Commun.}, 69:\penalty0 26--34, 2017.

\bibitem[Pintore et~al.(2021)Pintore, Almansa, and Schneider]{Pintore2021SliceNetDD}
Giovanni Pintore, Eva Almansa, and Jens Schneider.
\newblock Slicenet: deep dense depth estimation from a single indoor panorama using a slice-based representation.
\newblock \emph{2021 IEEE/CVF Conference on Computer Vision and Pattern Recognition (CVPR)}, pages 11531--11540, 2021.

\bibitem[Rey-Area et~al.(2022)Rey-Area, Yuan, and Richardt]{ReyArea2022360MonoDepthH3}
Manuel Rey-Area, Mingze Yuan, and Christian Richardt.
\newblock 360monodepth: High-resolution 360° monocular depth estimation.
\newblock \emph{2022 IEEE/CVF Conference on Computer Vision and Pattern Recognition (CVPR)}, pages 3752--3762, 2022.

\bibitem[Ronneberger et~al.(2015)Ronneberger, Fischer, and Brox]{Ronneberger2015UNetCN}
Olaf Ronneberger, Philipp Fischer, and Thomas Brox.
\newblock U-net: Convolutional networks for biomedical image segmentation.
\newblock In \emph{{MICCAI} {(3)}}, pages 234--241. Springer, 2015.

\bibitem[Shakerinava and Ravanbakhsh(2021)]{Shakerinava2021EquivariantNF}
Mehran Shakerinava and Siamak Ravanbakhsh.
\newblock Equivariant networks for pixelized spheres.
\newblock \emph{Proceedings of the 38th International Conference on Machine Learning, {ICML}}, abs/2106.06662, 2021.

\bibitem[Shen et~al.(2022)Shen, Lin, Liao, Nie, Zheng, and Zhao]{Shen2022PanoFormerPT}
Zhijie Shen, Chunyu Lin, Kang Liao, Lang Nie, Zishuo Zheng, and Yao Zhao.
\newblock Panoformer: Panorama transformer for indoor 360° depth estimation.
\newblock In \emph{European Conference on Computer Vision}, 2022.

\bibitem[Su and Grauman(2018)]{Su2018KernelTN}
Yu-Chuan Su and Kristen Grauman.
\newblock Kernel transformer networks for compact spherical convolution.
\newblock \emph{2019 IEEE/CVF Conference on Computer Vision and Pattern Recognition (CVPR)}, pages 9434--9443, 2018.

\bibitem[Sun et~al.(2020)Sun, Sun, and Chen]{Sun2020HoHoNet3I}
Cheng Sun, Min Sun, and Hwann-Tzong Chen.
\newblock Hohonet: 360 indoor holistic understanding with latent horizontal features.
\newblock \emph{2021 IEEE/CVF Conference on Computer Vision and Pattern Recognition (CVPR)}, pages 2573--2582, 2020.

\bibitem[Tan and Le(2019)]{Tan2019EfficientNetRM}
Mingxing Tan and Quoc~V. Le.
\newblock Efficientnet: Rethinking model scaling for convolutional neural networks.
\newblock In \emph{{ICML}}, pages 6105--6114. {PMLR}, 2019.

\bibitem[Tateno et~al.(2018)Tateno, Navab, and Tombari]{Tateno2018DistortionAwareCF}
Keisuke Tateno, Nassir Navab, and Federico Tombari.
\newblock Distortion-aware convolutional filters for dense prediction in panoramic images.
\newblock In \emph{European Conference on Computer Vision}, 2018.

\bibitem[Wang et~al.(2020)Wang, Yeh, Sun, Chiu, and Tsai]{Wang2020BiFuseM3}
Fu-En Wang, Yu-Hsuan Yeh, Min Sun, Wei-Chen Chiu, and Yi-Hsuan Tsai.
\newblock Bifuse: Monocular 360 depth estimation via bi-projection fusion.
\newblock \emph{2020 IEEE/CVF Conference on Computer Vision and Pattern Recognition (CVPR)}, pages 459--468, 2020.

\bibitem[Wang et~al.(2022{\natexlab{a}})Wang, Yeh, Tsai, Chiu, and Sun]{Wang2022BiFuseSA}
Fu-En Wang, Yu-Hsuan Yeh, Yi-Hsuan Tsai, Wei-Chen Chiu, and Min Sun.
\newblock Bifuse++: Self-supervised and efficient bi-projection fusion for 360° depth estimation.
\newblock \emph{IEEE Transactions on Pattern Analysis and Machine Intelligence}, 45:\penalty0 5448--5460, 2022{\natexlab{a}}.

\bibitem[Wang et~al.(2022{\natexlab{b}})Wang, Liang, Song, Li, and Wu]{Wang2022DABERTDA}
Sirui Wang, Di Liang, Jian Song, Yuntao Li, and Wei Wu.
\newblock {DABERT:} dual attention enhanced {BERT} for semantic matching.
\newblock In \emph{{COLING}}, pages 1645--1654. International Committee on Computational Linguistics, 2022{\natexlab{b}}.

\bibitem[Wang et~al.(2017)Wang, Girshick, Gupta, and He]{Wang2017NonlocalNN}
X. Wang, Ross~B. Girshick, Abhinav~Kumar Gupta, and Kaiming He.
\newblock Non-local neural networks.
\newblock \emph{2018 IEEE/CVF Conference on Computer Vision and Pattern Recognition}, pages 7794--7803, 2017.

\bibitem[yang Li et~al.(2022)yang Li, Guo, Yan, Huang, Duan, and Ren]{Li2022OmniFusion3M}
Yu yang Li, Yuliang Guo, Zhixin Yan, Xinyu Huang, Ye Duan, and Liu Ren.
\newblock Omnifusion: 360 monocular depth estimation via geometry-aware fusion.
\newblock \emph{2022 IEEE/CVF Conference on Computer Vision and Pattern Recognition (CVPR)}, pages 2791--2800, 2022.

\bibitem[Ye et~al.(2022)Ye, Shu, Li, Shi, Li, Wang, Tan, and Ding]{Ye2022Rope3DTR}
Xiaoqing Ye, Mao Shu, Hanyu Li, Yifeng Shi, Yingying Li, Guan-Sheng Wang, Xiao Tan, and Errui Ding.
\newblock Rope3d: The roadside perception dataset for autonomous driving and monocular 3d object detection task.
\newblock \emph{2022 IEEE/CVF Conference on Computer Vision and Pattern Recognition (CVPR)}, pages 21309--21318, 2022.

\bibitem[Yoon et~al.(2021)Yoon, Chung, Wang, and Yoon]{Yoon2021SphereSRI}
Youngho Yoon, Inchul Chung, Lin Wang, and Kuk-Jin Yoon.
\newblock Spheresr: $360^{\circ}$ image super-resolution with arbitrary projection via continuous spherical image representation.
\newblock \emph{2022 IEEE/CVF Conference on Computer Vision and Pattern Recognition (CVPR)}, pages 5667--5676, 2021.

\bibitem[Yu et~al.(2023)Yu, He, Jian, Feng, and Liu]{Yu2023PanelNetU3}
Haozheng Yu, Lu He, Bing Jian, Weiwei Feng, and Shanghua Liu.
\newblock Panelnet: Understanding 360 indoor environment via panel representation.
\newblock \emph{2023 IEEE/CVF Conference on Computer Vision and Pattern Recognition (CVPR)}, pages 878--887, 2023.

\bibitem[Yun et~al.(2023)Yun, Shin, Lee, Lee, and Rhee]{Yun2023EGformerEG}
Ilwi Yun, Chan-Yong Shin, Hyunku Lee, Hyuk-Jae Lee, and Chae-Eun Rhee.
\newblock Egformer: Equirectangular geometry-biased transformer for 360 depth estimation.
\newblock \emph{ArXiv}, abs/2304.07803, 2023.

\bibitem[Zhang et~al.(2019)Zhang, Liwicki, Smith, and Cipolla]{Zhang2019OrientationAwareSS}
Chao Zhang, Stephan Liwicki, William Smith, and Roberto Cipolla.
\newblock Orientation-aware semantic segmentation on icosahedron spheres.
\newblock \emph{2019 IEEE/CVF International Conference on Computer Vision (ICCV)}, pages 3532--3540, 2019.

\bibitem[Zhao et~al.(2020)Zhao, Jiang, Jia, Torr, and Koltun]{Zhao2020PointT}
Hengshuang Zhao, Li Jiang, Jiaya Jia, Philip H.~S. Torr, and Vladlen Koltun.
\newblock Point transformer.
\newblock \emph{2021 IEEE/CVF International Conference on Computer Vision (ICCV)}, pages 16239--16248, 2020.

\bibitem[Zheng et~al.(2019)Zheng, Zhang, Li, Tang, Gao, and Zhou]{Zheng2019Structured3DAL}
Jia Zheng, Junfei Zhang, Jing Li, Rui Tang, Shenghua Gao, and Zihan Zhou.
\newblock Structured3d: A large photo-realistic dataset for structured 3d modeling.
\newblock In \emph{European Conference on Computer Vision}, 2019.

\bibitem[Zheng et~al.(2023)Zheng, Zhu, Liu, Cao, Fu, and Wang]{Zheng2023BothSA}
Xu Zheng, Jinjing Zhu, Ye-Peng Liu, Zidong Cao, Chong Fu, and Lin Wang.
\newblock Both style and distortion matter: Dual-path unsupervised domain adaptation for panoramic semantic segmentation.
\newblock \emph{2023 IEEE/CVF Conference on Computer Vision and Pattern Recognition (CVPR)}, pages 1285--1295, 2023.

\bibitem[Zhuang et~al.(2021)Zhuang, Lu, Wang, Xiao, and Wang]{Zhuang2021ACDNetAC}
Chuanqing Zhuang, Zhengda Lu, Yiqun Wang, Jun Xiao, and Ying Wang.
\newblock Acdnet: Adaptively combined dilated convolution for monocular panorama depth estimation.
\newblock \emph{CoRR}, abs/2112.14440, 2021.

\bibitem[Zioulis et~al.(2018)Zioulis, Karakottas, Zarpalas, and Daras]{Zioulis2018OmniDepthDD}
Nikolaos Zioulis, Antonis Karakottas, Dimitrios Zarpalas, and Petros Daras.
\newblock Omnidepth: Dense depth estimation for indoors spherical panoramas.
\newblock In \emph{{ECCV} {(6)}}, pages 453--471. Springer, 2018.

\end{thebibliography}
}

\end{document}